\documentclass[conference]{IEEEtran}
\IEEEoverridecommandlockouts

\ifCLASSINFOpdf
\else
\fi
	\usepackage{graphicx}
	\usepackage{float}
	\usepackage{stfloats}
	\usepackage{amsmath}
	
	\usepackage{times}
	\usepackage{helvet}
	\usepackage{courier}
	\usepackage{amssymb}
	\usepackage{epsfig}
	\usepackage{booktabs}
	\usepackage{graphicx,subfigure}
	\usepackage{amsmath}
	\usepackage{amsthm}
	\usepackage[ruled, linesnumbered]{algorithm2e}
	\usepackage{nccbbb}
	\usepackage{url}
	\usepackage{color}
	\usepackage[utf8]{inputenc}
	\usepackage[small]{caption}
	\usepackage{array}
	\usepackage{colortbl}
	\usepackage{rotating}
	\usepackage{mathrsfs}
	\usepackage{multirow}
	\usepackage[backref]{hyperref} 
	\hypersetup{hypertex=true,
		colorlinks=true,
		linkcolor=blue,
		anchorcolor=blue,
		citecolor=blue}
	\usepackage{dsfont}
	\usepackage{enumerate}
	\usepackage{balance}
	
	\makeatletter
	
	\newcommand{\Rmnum}[1]{\expandafter\@slowromancap\romannumeral #1@}
	\makeatother
	\makeatletter
	\newcommand{\linebreakand}{%
	\end{@IEEEauthorhalign}
	\hfill\mbox{}\par
	\mbox{}\hfill\begin{@IEEEauthorhalign}
	}
	\makeatother
	
\begin{document}
		%
		\title{Word length-aware text spotting: Enhancing detection and recognition in dense text image}
		
		\author{\IEEEauthorblockN{Hao Wang}
			\IEEEauthorblockA{\textit{College of Computer Science and Engineering}\\
				\textit{Wuhan Institute of Technology} \\
				\textit{Hubei Key Laboratory of Intelligent Robot} \\
				Wuhan, China \\
				wangh@stu.wit.edu.cn}
			\and
			\IEEEauthorblockN{Huabing Zhou*}
			\IEEEauthorblockA{\textit{College of Computer Science and Engineering}\\
				\textit{Wuhan Institute of Technology} \\
				\textit{Hubei Key Laboratory of Intelligent Robot} \\
				Wuhan, China \\
				zhouhuabing@gmail.com}
			*Corresponding author
			\linebreakand 
			\IEEEauthorblockN{Yanduo Zhang}
			\IEEEauthorblockA{\textit{Hubei University of Arts and Science}\\
				Xiangyang, China \\
				zhangyanduo@hotmail.com}
			\and
			\IEEEauthorblockN{Tao Lu}
			\IEEEauthorblockA{\textit{College of Computer Science and Engineering}\\
				\textit{Wuhan Institute of Technology} \\
				\textit{Hubei Key Laboratory of Intelligent Robot} \\
				Wuhan, China \\
				lutxyl@gmail.com}
			\and
			\IEEEauthorblockN{Jiayi Ma}
			\IEEEauthorblockA{\textit{Electronic Information School}\\
				\textit{Wuhan University} \\
				Wuhan, China \\
				jyma2010@gmail.com}
		}

		\maketitle
		\begin{abstract}
			Scene text spotting is essential in various computer vision applications, enabling extracting and interpreting textual information from images. However, existing methods often neglect the spatial semantics of word images, leading to suboptimal detection recall rates for long and short words within long-tailed word length distributions that exist prominently in dense scenes. In this paper, we present \textbf{WordLenSpotter}, a novel \textbf{word} \textbf{len}gth-aware \textbf{spotter} for scene text image detection and recognition, improving the spotting capabilities for long and short words, particularly in the tail data of dense text images. We first design an image encoder equipped with a dilated convolutional fusion module to integrate multiscale text image features effectively. Then, leveraging the Transformer framework, we synergistically optimize text detection and recognition accuracy after iteratively refining text region image features using the word length prior. Specially, we design a Spatial Length Predictor module (SLP) using character count prior tailored to different word lengths to constrain the regions of interest effectively. Furthermore, we introduce a specialized word Length-aware Segmentation (LenSeg) proposal head, enhancing the network's capacity to capture the distinctive features of long and short terms within categories characterized by long-tailed distributions. Comprehensive experiments on public datasets and our dense text spotting dataset DSTD$1500$ demonstrate the superiority of our proposed methods, particularly in dense text image detection and recognition tasks involving long-tailed word length distributions encompassing a range of long and short words.
	\end{abstract}
	

	\IEEEpeerreviewmaketitle

	\section{Introduction}
	
	The purpose of scene text spotting is to obtain text information from natural scene images or videos through detection and recognition, which is of great value to many real-world applications such as image retrieval, automatic driving, and assistance for blind people. With the impressive representational capacity of deep neural networks, scene text spotting has witnessed significant advancements in recent years. However, the long-tailed distribution phenomenon characterized by word length poses a significant challenge to text-spotting tasks, especially in dense scenes with large data samples. The missed detection of long and short words becomes a significant hurdle to overcome in dense scene text spotting.
	
	The traditional scene text spotting methods, such as Tian et al. \cite{tian2016detecting} and Liao et al.  \cite{liao2018textboxes++}, treated detection and recognition as separate tasks and follow a pipeline approach to handle text detection and recognition independently. Typically, a pre-trained text detector \cite{zhou2017east} is employed to generate text proposals, which are then passed to a text recognizer \cite{shi2016end} for obtaining the final results. It's essential to have accurate text detection as it significantly impacts the recognition results, and inaccurate detection can lead to performance issues due to error accumulation. To address this issue, researchers proposed end-to-end methods \cite{li2017towards,liu2018fots} that integrate text detection and recognition into a unified network. In these approaches, the detection branch extracts text instances, while the recognition branch predicts sequence labels for each text instance. By training the model in an end-to-end manner, the detection and recognition components assist each other, resulting in significant improvements in the performance of scene text spotting.
	
\begin{figure*}[tbp]
		\centering 
		\includegraphics[width=\linewidth]{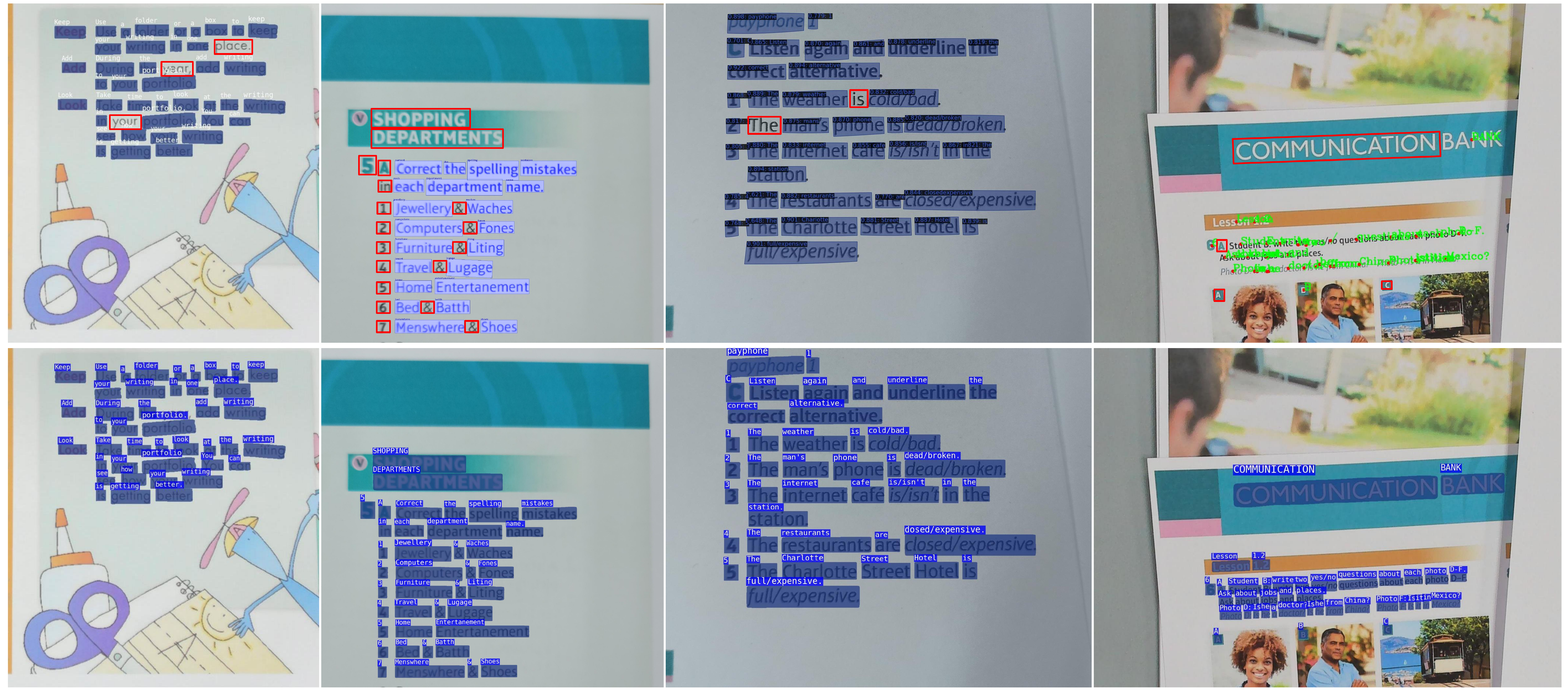}
		\caption{The recent state-of-the-art methods in the visualization of recognition results in dense scenes,  the first row from left to right is SwinTextSpotter \cite{huang2022swintextspotter}, PAN++ \cite{wang2021pan++}, TESTR \cite{zhang2022text}, and SPTS \cite{peng2022spts}, all have the phenomenon for missed detection of long and short words. The second row is the performance of our method. Red rectangular boxes indicate missed detections.}
		\label{miss_detection_figure}
	\end{figure*}

	However, the majority of existing scene text spotting methods, such as Wang et al. \cite{wang2021pan++}, Huang et al. \cite{huang2022swintextspotter}, Zhang et al. \cite{zhang2022text}, and Peng et al. \cite{peng2022spts}, fail to effectively handle dense text images, as illustrated in Fig. \ref{miss_detection_figure}. This limitation arises from the severe class imbalance issue in real-world data when sufficient samples are available. Text spotting data is divided by word length; the majority of labels (tail) are associated with a limited number of instances, while a few categories (head) dominate the samples. These methods often overlook the spatial semantics of words in images and do not specifically design training for the tail data, resulting in unsatisfactory recall rates for detecting long and short words within the long-tailed word length distribution.
	
	To address these limitations, we present a comprehensive and end-to-end trainable word length-aware spotter framework called WordLenSpotter. Firstly, our designed dilated convolutional fusion module can successfully integrate text image features at multiple scales. Then, we utilize the Transformer \cite{vaswani2017attention} framework to synergistically optimize the accuracy of text detection and recognition after iteratively refining text image Region of Interest (RoI) features using word length prior. Specifically, considering that dense text images have a more pronounced word length distribution, we introduce Spatial Length Predictor for different word lengths to constrain the regions of interest effectively. Furthermore, to enhance the capacity of our network to capture distinctive features of long and short terms within long-tailed categories, we designed a specialized word Length-aware Segmentation proposal head. Finally, we constructed a dense scene text dataset of various resolutions from daily reading scenarios called DSTD$1500$, where a single image contains more than $50$ text instances on average. We evaluate its performance on the multi-oriented dataset ICDAR$2015$ \cite{karatzas2015icdar}, arbitrarily shaped dataset Total-Text \cite{ch2020total} and CTW$1500$ \cite{liu2019curved}, and our dense scene text dataset DSTD$1500$. The experimental results demonstrate that WordLenSpotter outperforms state-of-the-art methods in terms of recall and precision for text spotting in dense scenes. Its exceptional proficiency in handling short and long text instances with long-tailed word length distributions makes it highly suitable for effectively processing images containing dense text. The significant contributions of our work can be summarized as follows:
	
	\begin{itemize}
		\item To our knowledge, we propose for the first time a specialized word Length-aware Segmentation (LenSeg) Head to enhance the capacity of the network to capture unique features of long and short words in dense text images with long-tailed word length distribution.
		\item Our proposed WordLenSpotter utilizes Swin Transformer and BiFPN as image encoder, employing dilated convolution to expand the receptive field and effectively fuse multiscale features, enhancing the effectiveness of the method.
		\item For the first time, we consider the spatial semantics of text in images and effectively constrain the region of interest using the proposed Spatial Length Predictor (SLP), which allows for the efficient incorporation of text priors with different word lengths into the collaborative architecture of text detection and recognition.
		\item We construct a dense scene text dataset (DSTD$1500$) under conventional paper reading scenarios to evaluate the performance of our WordLenSpotter network. The text instances in these images are densely distributed, with horizontal, multi-oriented, and curved shapes. 
		\item Extensive experiments show that our WordLenSpotter network can achieve state-of-the-art text spotting performance on our dense text scene dataset DSTD$1500$. Meanwhile, our WordLenSpotter network also offers strong competitiveness on public datasets.
	\end{itemize}
	
	The remainder of this paper is organized as follows. In Section \ref{Section_2}, we describe the related work of our study. In Section \ref{Section_3}, we introduce the definition of priors adopted in WordLenSpotter and the two innovative branch tasks, and make a statement about label generation. In Section \ref{Section_4}, we briefly introduce the training datasets adopted and analyzed the effectiveness of WordLenSpotter through experimental results on different datasets. Section \ref{Section_5} shows our conclusions.

	\section{Related work}\label{Section_2}
	
	\subsection{Scene text detection}
	
	Scene text detection is an important component of scene text spotting. A text extraction system iscalled PhotoOCR \cite{bissacco2013photoocr} was established using deep neural networks and HOG features, and based on character classification methods. Tian et al. \cite{tian2016detecting} utilized a vertical anchoring mechanism to predict category scores for suggested text positions. Inspired by single-stage object detection methods, Liao et al. \cite{liao2017textboxes} employed a fully convolutional network to achieve word-level horizontal text detection. In the same year, an efficient scene text detection method called EAST \cite{zhou2017east} was proposed, which achieved multi-oriented text detection using rotating boxes. Shortly thereafter, Liao et al. \cite{liao2018textboxes++} introduced quadrilateral offset prediction to achieve multi-oriented quadrilateral text detection. Baek et al. \cite{baek2019character} achieved text detection of arbitrary shapes through character perception, but character-level annotation was expensive and the model required cumbersome post-processing. Wang et al. \cite{wang2019arbitrary} successfully employed paired boundary point of the order prediction adaptive number to achieve text detection of any shape, but it is only limited to sequence prediction of RNNs. Recently, Yu et al. \cite{yu2023turning} used pre-trained models to compare text image pairs, reducing data annotation requirements for text detection and enabling training with small samples, but at the cost of reduced inference efficiency. Unlike them, we leverage the word length prior in dense text images, enabling effective detection of a large number of text instances with arbitrary shapes.
	
	\subsection{Scene text recognition}
	
	Scene text recognition has evolved from regular text recognition to tackling more challenging irregular text. Early scene text recognition methods \cite{bissacco2013photoocr,yao2014strokelets} all followed a bottom-up approach. These methods involved region segmentation, character classification, and word assembly, but were limited by expensive character-level annotations. Su et al. \cite{su2017accurate} pioneered the introduction of recurrent neural networks into scene text recognition and proposed a scene text recognition system based on HOG features and recurrent neural networks. Shi et al. \cite{shi2016end} proposed a scene text recognition method called CRNN. This method integrated feature extraction, sequence modeling, and transcription into a unified framework to achieve regular text recognition. Meanwhile, Lee et al. \cite{lee2016recursive} successfully incorporated the attention-based sequence-to-sequence model into scene text recognition. This advancement in text recognition methods has led to increased maturity and a shift towards recognizing text in irregular scenes. Luo et al. \cite{luo2019moran} corrected irregular text by predicting the offset of each part of the input image. Liu et al. \cite{liu2023towards} introduced an open-set text recognition task that enhances the practical applicability of text recognition. In this paper, considering the spatial semantics of dense text images, our text recognition module incorporates spatial attention to enhance the prediction of text sequences with arbitrary shapes.
	
	\subsection{Scene text spotting}
	
	In recent years, the end-to-end scene text spotting method has made significant progress in achieving optimization for scene text detection and recognition systems. Li et al. \cite{li2017towards} introduced variable size RoI Pooling to effectively extract text proposal features and achieved an end-to-end regular quadrilateral scene text spotting method. However, the quantization process of RoI Pooling led to a misalignment between the RoI and extracted features. Liu et al. \cite{liu2018fots} achieved multi-oriented text detection and recognition using RoI-Rotate and Text Align-Sampling, but it cannot effectively spot arbitrarily shaped text. The Mask Textspotter series \cite{lyu2018mask,liao2021mask,liao2020mask}  pioneered the use of RoIAlign to address the misalignment issue between RoI and feature extraction, successfully achieving recognition of arbitrary-shaped text. Feng et al. \cite{feng2019textdragon} adopted RoISlide to predict quadrilaterals on the centerline, achieving end-to-end word-level text spotting of arbitrary shapes. The Text Perceptron proposed by \cite{qiao2020text} was based on a semantic segmentation method that uses TPS to transform the detected text regions into regular shapes. Ronen et al. \cite{ronen2022glass} improved recognition performance with a global-to-local attention module and Rotated-RoIAlign. In order to avoid the complex design caused by RoI operations, both \cite{wang2021pgnet} and \cite{qiao2021mango} proposed character-level arbitrary shape text detection and recognition methods that do not require RoI operations. The ABCNet series \cite{liu2020abcnet,liu2021abcnet} proposed by Liu et al. utilizes BezierAlign to convert arbitrarily shaped text into regular text. Wang et al. \cite{wang2021pan++} designed a lightweight scene text spotting network based on text kernels and pixel clustering. Recently, in order to reduce the annotation cost of end-to-end scene text spotting, Peng et al. \cite{peng2022spts} attempted a scene text spotting method based on weak position information supervision. With the introduction of the Pix-to-Seq \cite{chen2021pix2seq} paradigm and the popularity of DETR \cite{carion2020end} and its variants, groundbreaking work \cite{huang2022swintextspotter,zhang2022text,kittenplon2022towards} in the field of scene text spotting had shown good performance in the set prediction methods that do not require NMS post-processing. Inspired by these works, we focus on the distinctive word length distribution characteristics in dense text images and leverage a collaborative optimized architecture combining scene text detection and recognition to achieve dense text spotting.
	
	\section{Methodology}\label{Section_3}
	
\begin{figure*}[t]
		\centering 
		\includegraphics[width=\linewidth]{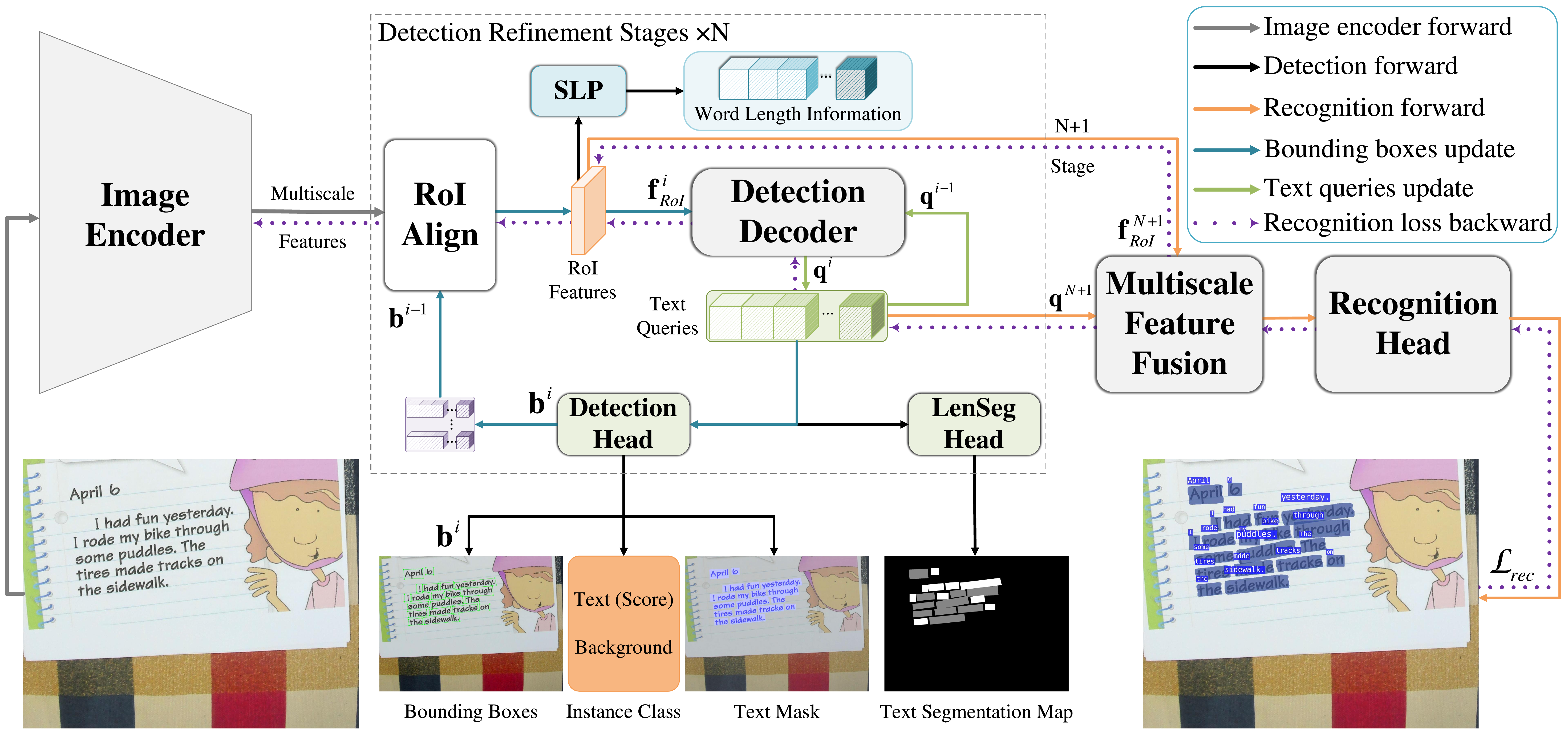}
		\caption{The WordLenSpotter architecture. It utilizes $N$ detection refinement stages followed by a joint stage to iteratively refine its outputs. In each detection stage, RoIAlign aligns multiscale features based on the updated text bounding boxes (by \textcolor[RGB]{49,133,156}{$\longrightarrow$}), producing RoI features. The Detection Decoder then updates the text queries (by \textcolor[RGB]{157,187,97}{$\longrightarrow$}) using these RoI features. In the $(N+1)$-th stage, higher-resolution RoI features are used for recognition. Multiscale Feature Fusion combines these features with the advanced text queries from the $(N+1)$-th stage. Recognition Head performs text recognition on the refined detection results. Throughout the $N$ detection refinement stages, SLP predicts word length information, while the Detection Head refines text bounding boxes and predicts instance class and text mask. The LenSeg Head is responsible for word length-aware segmentation using the text segmentation map. The symbol (\textit{e.g.}, $\mathbf{b}^{i-1}, \mathbf{q}^{i-1},\mathbf{f}^i_{RoI},  \mathbf{q}^{N+1}$ and $\mathbf{f}^{N+1}_{RoI}$) at the end of the arrow represents input of the module, while the symbol (\textit{e.g.}, $\mathbf{b}^{i}$ and $\mathbf{q}^{i}$) at the begin of the arrow represents output of the module.}
		\label{WordLenSpotter_figure}
	\end{figure*}

	The WordLenSpotter architecture (shown in Fig. \ref{WordLenSpotter_figure}) iteratively refines its output through $N$ detection stages followed by a joint stage. Initially, image features are extracted at various scales using the image encoder. In each detection stage ($i$-th stage, $i = 1,\cdots, N$), RoIAlign is employed to generate RoI features $\mathbf{f}^i_{RoI}$ from the text bounding boxes $\mathbf{b}^{i-1}$ and image features. The features $\mathbf{f}^i_{RoI}$, along with text queries $\mathbf{q}^{i-1}$, are fed into a transformer decoder to refine the queries into $\mathbf{q}^i$. By applying the detection head to $\mathbf{q}^i$, the text bounding boxes $\mathbf{b}^{i}$ are improved, resulting in finer-grained RoI features $\mathbf{f}^{i+1}_{RoI}$. The detection stage also outputs the detected instance class and the text mask required by the recognition head \cite{liao2021mask} through a detection head composed of simple linear layers.
	
	To consider spatial semantics in dense text images, we design a Spatial Length Predictor (SLP), which extracts text aspect ratio and character quantity information from $\mathbf{f}^i_{RoI}$ and leverages word length information to enforce the RoI features to capture word spatial semantics. Additionally, the word Length-aware Segmentation (LenSeg) Head is designed specifically for the tail data of unbalanced word length samples. It employs a straightforward proposal network to generate a text segmentation map that indicates the location information of long and short words based on $\mathbf{q}^i$.
	
	In the $(N+1)$-th stage, the resolution of RoI features is twice that of the detection stage. As recognition features, $\mathbf{f}^{N+1}_{RoI}$ are downsampled to provide richer recognition information. The recognizer utilizes the detection feature $\mathbf{q}^{N+1}$ from the $(N+1)$-th stage and the recognition feature for sequence prediction using the multiscale fusion module \cite{huang2022swintextspotter}. This end-to-end trainable framework optimizes detection and recognition results in a synergistic manner.
	
	\subsection{Overview}
	
	\subsubsection{Preliminary}
	
	The text detection and recognition collaboration architecture proposed in SwinTextSpotter \cite{huang2022swintextspotter} is suitable for text spotting in conventional scenes. It regards the text detection task as a set prediction problem and uses a set of learnable proposed boxes and text queries to learn the position of text in the image and semantic information within the image, respectively. The detection decoder can calculate the correlation between RoI features and text queries. The architecture uses the same decoder with a dynamic head to decode the same set of text queries during the detection and recognition stages, and multiscale fusion is performed on the recognition features obtained through RoIAlign and text queries containing rich detection features. Therefore, the recognizer can optimize the detection results through the loss gradient backflow predicted by the character sequence. However, it is important to note that when applied solely to dense text images, there may still be instances where the detection process misses certain text instances. Drawing inspiration from SwinTextSpotter \cite{huang2022swintextspotter}, we explored and studied the task of text spotting in dense scenes.
	
	\subsubsection{The image encoder of WordLenSpotter}\label{Section_3_1_2}
	
	\begin{figure}[t]
		\centering 
		\includegraphics[width=\linewidth]{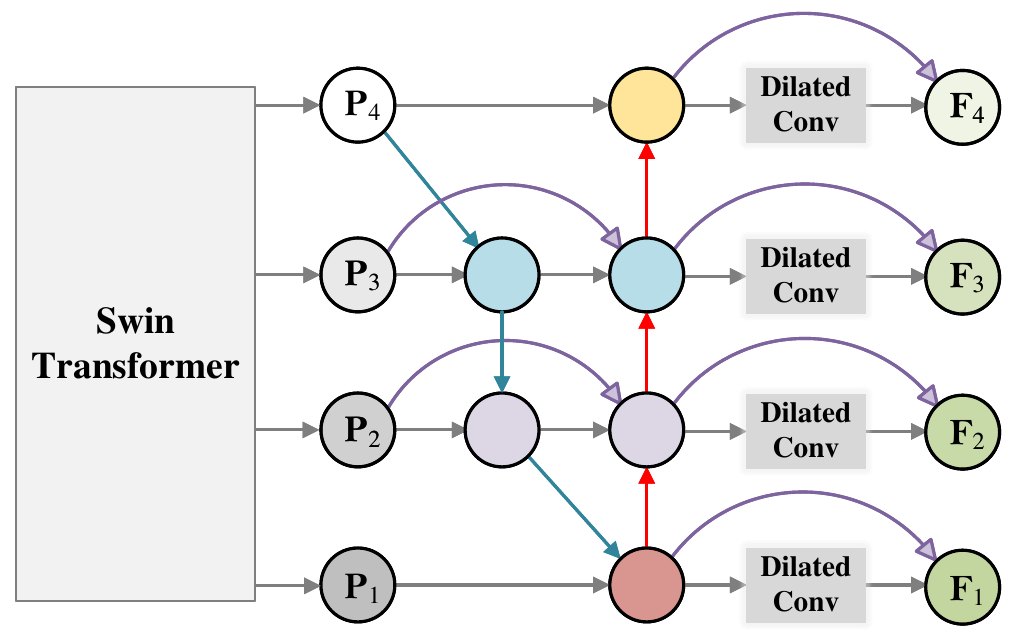}
		\caption{Details of the image encoder network in WordLenSpotter. The image is extracted by Swin Transformer \cite{liu2021swin} to obtain four feature maps with different scales and dimensions, $\mathbf{P}_1$, $\mathbf{P}_2$, $\mathbf{P}_3$, and $\mathbf{P}_4$. These feature maps are then processed through BiFPN \cite{tan2020efficientdet} and dilated convolution to obtain fused features with different scales but with the same dimensions $\mathbf{F}_1$, $\mathbf{F}_2$, $\mathbf{F}_3$, and $\mathbf{F}_4$. \textcolor[RGB]{127,127,127}{$\longrightarrow$} in the figure represents the inflow and outflow of data, \textcolor[RGB]{49,133,156}{$\longrightarrow$} represents upsampling, \textcolor[RGB]{255,0,0}{$\longrightarrow$} represents downsampling,  \textcolor[RGB]{127,101,159}{$\longrightarrow$} represents residual shorting structure, and the circles represent different feature maps.}
		\label{bifpn_figure}
	\end{figure}
	
	The image encoder takes the original image as input and extracts advanced features for subsequent detection and recognition decoders. Due to the strong similarity between different texts in the text spotting task of dense scenes, it is crucial to globally model dense scene images to construct relationships between different texts. The image encoder of WordLenSpotter is shown in Fig. \ref{bifpn_figure}. Our proposed WordLenSpotter adopt Swin Transformer \cite{liu2021swin} based on shift window to extract low-level representation and high-level semantic information in scene images from the simple to the deep. In order to more fully integrate the shallow and deep features extracted by the Swin Transformer without increasing too much cost, we adopt an efficient bidirectional cross-scale connection and weighted feature fusion network (BiFPN \cite{tan2020efficientdet}) with only four inputs. Considering that there are large or small gaps between text instances in the scene image, the network needs a larger receptive field to distinguish different text instances. Therefore, we employ the dilated convolution with residual short circuit structure to process the features after the second weighted fusion in BiFPN. Expanded BiFPN achieves multiscale feature fusion and expands the receptive field, enabling better text relationship modeling.
	
	\subsection{Spatial length predictor} \label{Section_3_2}
	
	\subsubsection{The relationship between character count and aspect ratio} \label{Section_3_2_1}
	
	In dense scene text spotting, most query-based methods often underperform due to the challenges posed by a high number of text instances and their random distribution. Matching text query results with ground truth becomes difficult. To address this issue, we consider the spatial semantics of text images and introduce a different word length prior information prediction branch that combines the strong correlation between text aspect ratio and character count. The aspect ratio, denoted as the ratio of the horizontal length ($l_{hs}$) to the vertical length ($l_{vs}$) of the closed graph formed by the boundary points of the text instance, exhibits a strong correlation with the number of characters in scene text images. To avoid naming ambiguity caused by character arrangement direction, we specify that the horizontal edge is the edge that maintains the same direction as the character arrangement direction, which is also in line with the habit of data annotation.
	
	On the premise of the above statement, we believe that from the training sample, Image contains $Y$ text instances, with the aspect ratio of each text instance denoted as $R_i$ and the number of characters contained in that text instance denoted as $N_i$. For any single text instance, there will always be: 
	\begin{equation}
		R_i\propto N_i, i\in\{1,2,\ldots ,Y\}.
	\end{equation}
	
	There is a positive correlation between the number of characters and the aspect ratio, meaning that as the number of characters increases, the aspect ratio tends to be greater. The prediction of text instance aspect ratio and character count can be seen as a simple regression task. Owing to their strong correlation, network parameter sharing can be achieved in the SLP. More details on aspect ratio and character count will be explained in Sec. \ref{Section_3_4_1}.
	
	\subsubsection{Text word length information prediction} \label{Section_3_2_2}
	
	Text word length prediction relies on WordLenSpotter to learn the relative length and character count information of text instances from the coordinates of their text bounding boxes. By enabling text queries to capture the spatial semantics of text images, we utilize the SLP to accurately predict the aspect ratio and character count of the text. 
	
	During the detection phase, the image features and detection results predict-boxes are processed through RoIAlign to obtain RoI features $\mathbf{f}_{RoI}\in \mathbb{R}^{B\times C\times7\times7}$, which contain the position information of the bounding boxes of the text instance, where $B$ is the product of the batch size and the number of text queries $Q_{num}$, and $C$ is the number of feature channels. Note that $B$ is greater than the number of text instances $Y$, and the detection phase needs to match the first $Y$ best predictions as output. Naturally, the prediction branch for text aspect ratio and text character count takes RoI features $\mathbf{f}_{RoI}$ from the detection stage as input. We have taken into account the dimensional characteristics of $\mathbf{f}_{RoI}$ in our SLP design. We first adopt a convolutional kernel of size $3\times3$ to learn the aspect ratio and character quantity information in RoI features: 
	\begin{equation}
		\mathbf{f}_1=\mathcal{F}_{\text{CBR}}(\mathbf{f}_{RoI}),
	\end{equation}
	where $\mathcal{F}_{\text{CBR}}$ represents a sampling process consisting of convolution, batch normalization, and the ReLU activation function. To address the potential feature loss resulting from downsampling, we employ convolutional kernel processing with the same kernel size as before, rather than pooling, in both downsampling processes $\mathcal{E}_{\text{down1}}$ and $\mathcal{E}_{\text{down2}}$. The step size is adjusted to $2$ and the padding is set to $0$. This approach helps mitigate the impact of downsampling on feature preservation. 
	\begin{equation}
		\mathbf{f}_2=\mathcal{E}_{\text{down2}}(\mathcal{E}_{\text{down1}}(\mathbf{f}_1)).
	\end{equation}
	
	The sampled features $\mathbf{f}_2 \in \mathbb{R}^{B\times C\times1\times1}$ are predicted through the linear layer of the prediction branch to obtain the text aspect ratio prediction results $\mathbf{R}_{p}\in \mathbb{R}^{B\times1\times1}$ and the text character number prediction results $\mathbf{N}_{p}\in\mathbb{R}^{B\times1\times1}$: 
	\begin{equation}
		(\mathbf{R}_{p}, \mathbf{N}_{p})= \omega \mathbf{f}_2+\beta,
	\end{equation}
	where $\omega$ and $\beta$ are the learnable parameter for the linear layer. The predicted set of text word length prior information $\mathbf{C}_{prior}$ can be represented as the concatenation of $\mathbf{R}_{p}$ and $\mathbf{N}_{p}$.
	
	We employ self-generated labels to constrain the prior information regression prediction tasks of text word length. And we incorporate the cost of prior text prediction directly into the bipartite matching calculation process of the detection results. By incorporating this cost, we ensure that the prior information influences the matching process. Therefore, text queries need to learn more accurate text position and character sequence information to achieve lower matching costs and achieve more accurate prediction in the detection stage.
	
	\subsection{Word length-aware segmentation head} \label{Section_3_3}
	
	\subsubsection{Definition of extreme word length text} \label{Section_3_3_1}
	
	We analyze the objective reasons for the poor performance of query-based methods in dense text spotting, which include the miss of many instances of short and long text, resulting in lower recall rates. This can be attributed to the increasing number of rare occurrences of short and long word samples as the overall text quantity increases in typical scenes. Perceiving these extreme word length texts through text queries poses challenges, yet their accurate spotting remains meaningful. To improve the precision and recall of text spotting in dense scenes, we propose to incorporate word Length-aware Segmentation proposal head in the collaborative architecture of text detection and recognition to enhance learning of short and long word features. 
	
\begin{figure}[t]
		\centering
		\includegraphics[width=\linewidth]{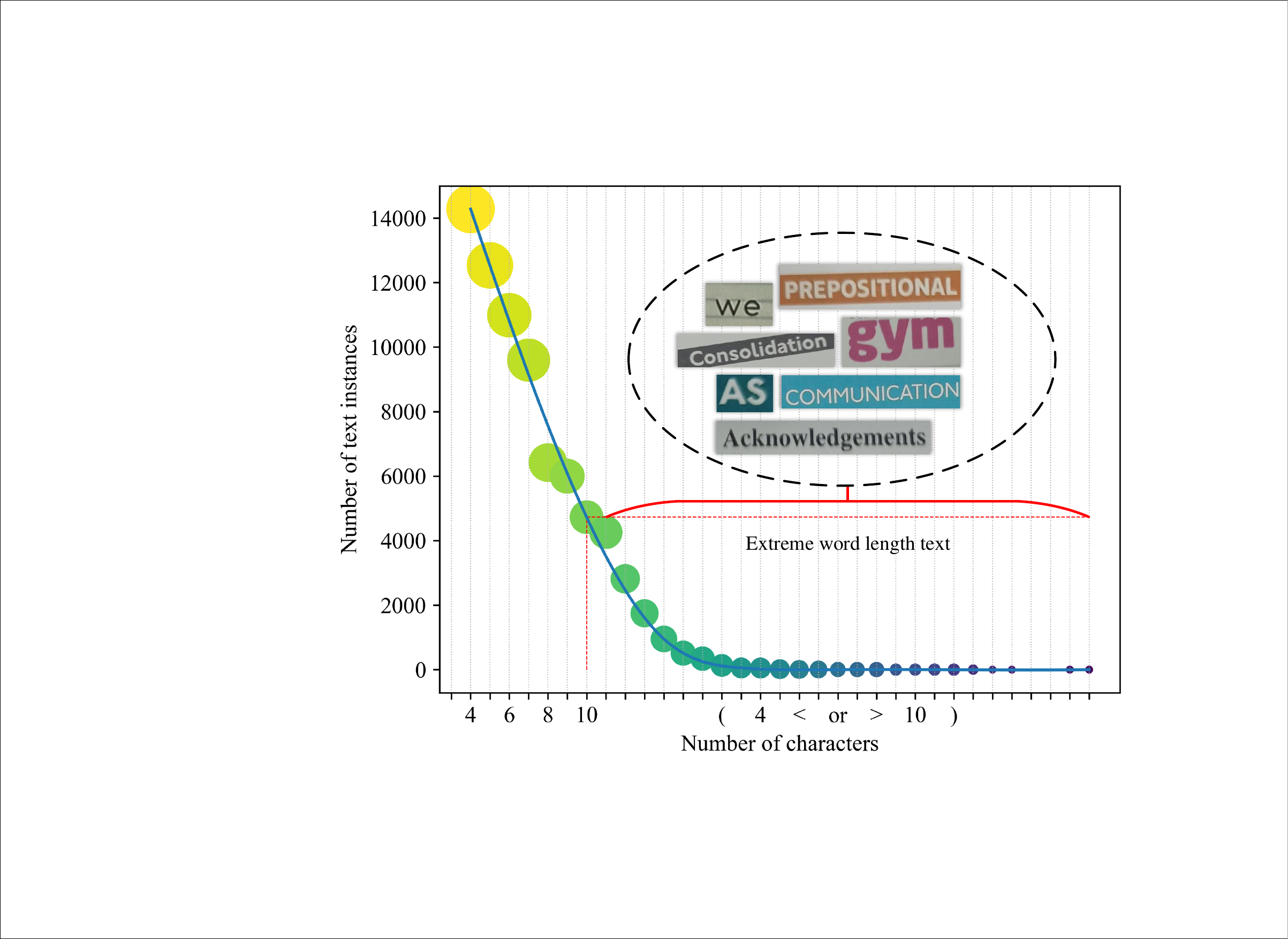}
		\caption{Character count statistics for DSTD$1500$ text instances. The distribution of text word lengths follows a long-tailed pattern, with the majority of words ranging from $4$ to $10$ characters, representing the most common word lengths. There are relatively few words with lengths below $4$ or above $10$, indicating instances of extreme word length text.}
		\label{dstd1500_data_analysis_b_figure}
	\end{figure}

	We use DSTD$1500$ as an example to analyze the relationship between text instances and the number of characters, as shown in Fig. \ref{dstd1500_data_analysis_b_figure}. Especially when an adequate number of text samples are available, the long-tailed distribution of word lengths is a common phenomenon in text data. It signifies that the majority of textual vocabulary falls within the range of $4$ to $10$ characters, which represents the most frequent and typical word lengths. On the other hand, the tail samples in the data distribution are words with shorter or longer lengths, which belong to lower frequencies or extreme situations. The reasons behind this distribution pattern may lie in the fact that common words and phrases in natural language tend to have lengths ranging from $4$ to $10$ characters, while shorter or longer word lengths occur less frequently. Additionally, when writing text, people tend to use words of typical lengths, further reinforcing this distribution pattern.
	
	Specifically, if a text instance contains a number of characters $N_i\in[4, 10]$, it is called a regular text instance. if a text instance contains a number of characters $N_i\in(0, 3]\cup[10, N_{max}]$, it is called an extreme word length text instance, where $N_{max}$ is the maximum character limit set based on dataset characteristics, and the number of characters in all text instances should not exceed $N_{max}$.
	
	\subsubsection{Word length-aware segmentation} \label{Section_3_3_2}
	
	We divide scene image content into three categories: background, regular text, and extreme word-length text. In order to give text queries more attention to extreme word-length texts that are more difficult to spot, and enhance the capacity of our network to capture unique features of long and short words in the long-tailed distribution, we adopt a specialized word Length-aware Segmentation Head called LenSeg Head. During the detection phase, we use a simple proposal head to generate a text segmentation map $\mathbf{C}_{map}$ of size $W\times H$ as an additional output from text queries. Additionally, we utilize the self-generated text segmentation map label to guide the generation of the text segmentation map, so that text queries can effectively learn the absolute position of the image region where extreme word-length texts are located.
	
	\subsection{Label generation} \label{Section_3_4}
	
	\subsubsection{Text word length information label} \label{Section_3_4_1}
	
	The labels required for SLP have similar generation process in different data annotations. Given an image containing $Y$ text instances, the text word length information label $\mathbf{G}_{prior}$ is composed of the aspect ratio $R_i$ of all text instances and the number of characters $N_i$, described as:
	\begin{equation}
		\mathbf{G}_{prior}=\{R_i,N_i \}_{i=1}^Y.
	\end{equation}
	
	In fact, $N_i$ can be obtained by calculating the length of transcription labels in data annotation. For a text instance in image, whose transcription annotation \cite{karatzas2015icdar} is $\mathbf{T}^{char}_i$, then: 
	\begin{equation}
		{N}_i=len(\mathbf{T}^{char}_i),
	\end{equation}
	where $len()$ is the character length calculation function.

	\begin{figure*}[tbp]
		\centering
		\begin{minipage}[t]{0.36\linewidth}
			\centering
			\includegraphics[width=\textwidth]{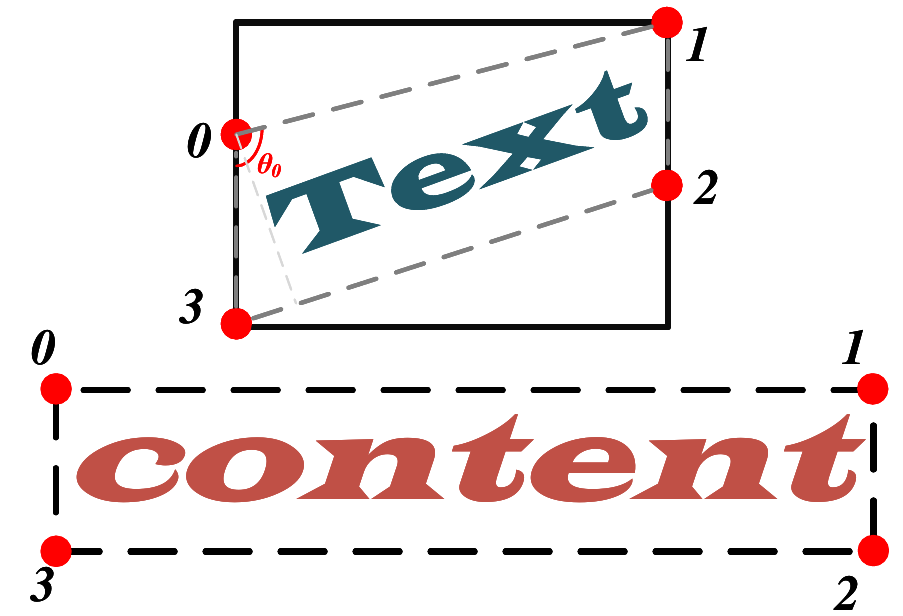}
			\centerline{(a) Horizontal and multi-oriented text}
		\end{minipage}
		\begin{minipage}[t]{0.36\linewidth}
			\centering
			\includegraphics[width=\textwidth]{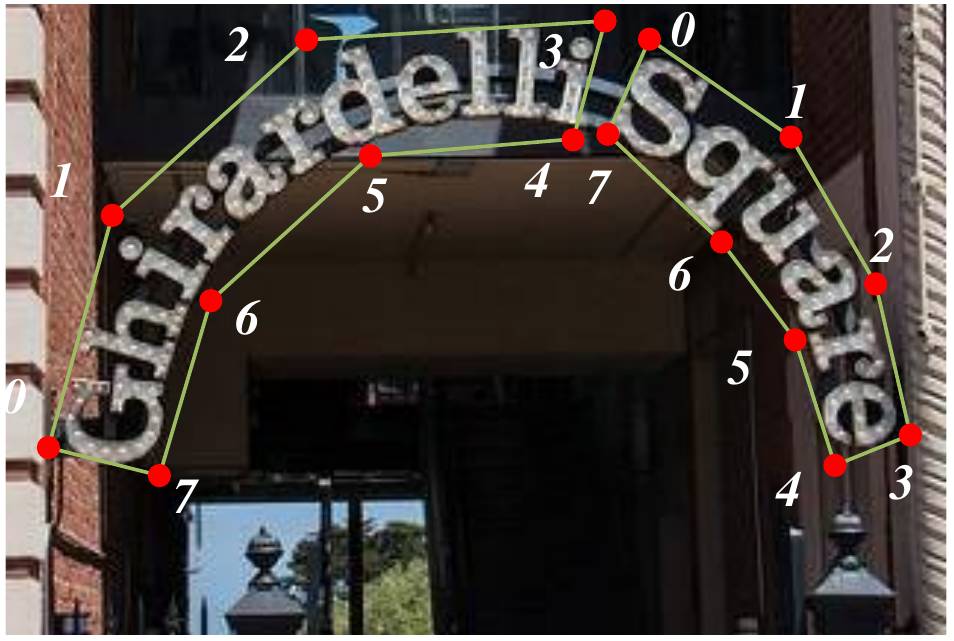}
			\centerline{(b) Curved and long text}
		\end{minipage}
		\caption{Explanation of the text word length prior count label generation. Each text instance contains several annotation coordinate points, numbered starting from $0$ in the upper left corner of the text instance.}
		\label{aspect_ratio_define_figure}
	\end{figure*}
	
	The aspect ratio calculation relies on the absolute position coordinates annotated by the text instance. Owing to the different number of coordinate points in different datasets, we divide the calculation of aspect ratio into two categories: horizontal text or multi-oriented text, which contains four coordinate points. The other type is curved or long text, containing $2n$ coordinate points, where $n\in \{3,4,5,6,7,8\}$. For the convenience of calculation, as shown in Fig. \ref{aspect_ratio_define_figure}, we mark the coordinate points clockwise starting from the top left corner of the text instance according to the labeling convention.
	
	Next, calculate the aspect ratio $R_i$ of a text instance. For horizontal or multi-oriented text, as shown in Fig. \ref{aspect_ratio_define_figure}(a), according to the definition of horizontal edges in Sec. \ref{Section_3_2_1}: 
	\begin{equation}
		l_{hs}=\frac{1}{2}\sum_{j=0}^1 |P_{2j+1}-P_{2j}|,
		\label{lhs_eq}
	\end{equation}
	where $l_{hs}$ is the horizontal length of the text instance, $P$ is the coordinate point marked by the text instance, and $|\cdot|$ is the euclidean  distance between two points. The horizontal edge length $l_{hs}$ is the average length of the two edges in the quadrilateral formed by the annotation points of the text instance that are aligned in the same direction as the text characters. 
	
	For a triangle composed of any three of the four coordinate points in the image plane, the Law of Cosines is satisfied: 
	\begin{equation}
		cos\theta=\frac{a^2+b^2-c^2}{2ab},
	\end{equation}
	where $a$, $b$, and $c$ are three edges composed of any three points. Calculate the four angles $\theta_j$ formed by the vertical and horizontal edges based on the coordinate values of the four points. Thus, it can be calculated that:  
	\begin{equation}
		l_{vs}=\frac{1}{4}\sum_{j=0}^3|P_j-P_{3-j}|sin\theta_j,
		\label{lvs_eq}
	\end{equation}
	where $l_{vs}$ is the vertical edge length of the text instance, $\theta_j$ is the inner angle of the quadrilateral at coordinate point $P_j$. The length of the longitudinal edge $l_{vs}$ is the average length calculated for the edges perpendicular to the arrangement direction of text characters in the quadrilateral formed by annotation points of text instances. The ratio of the length of the horizontal edge to the length of the vertical edge is denoted as the aspect ratio $R_i$:  
	\begin{equation}
		R_i=\frac{l_{hs}}{l_{vs}}.
		\label{R_eq}
	\end{equation}
	
	For curved or long text with $2n$ points and $n\in\{3,4,5,6,7,8\}$, all coordinate points are divided into $n-1$ rectangles according to the rule of grouping adjacent four coordinate points into a quadrilateral. For example, if each text instance in Fig. \ref{aspect_ratio_define_figure}(b) contains $8$ points, each text instance is divided into three quadrangles. Then, apply the Eqs. (\ref{lhs_eq}), (\ref{lvs_eq}), and (\ref{R_eq}) to calculate the horizontal and vertical lengths of $n-1$ quadrilaterals, resulting in the values $l_{hs}^k$ and $l_{vs}^k$. The ratio of the sum is denoted as the aspect ratio of the curved or long text:  
	\begin{equation}
		R_i=\frac{\sum_{k=1}^{n-1}l_{hs}^k}{\frac{1}{n-1}\sum_{k=1}^{n-1}l_{vs}^k}.
	\end{equation}
	
	We consider the differences in sample data and normalize the label $\mathbf{G}_{prior}\in \mathbb{R}^{Y\times1\times2}$ required for the SLP, where $Y$ represents the number of text instances in this image.
	
	\subsubsection{Text segmentation map label} \label{Section_3_4_2}
	
\begin{figure*}[t]
		\centering 
		\includegraphics[width=\linewidth]{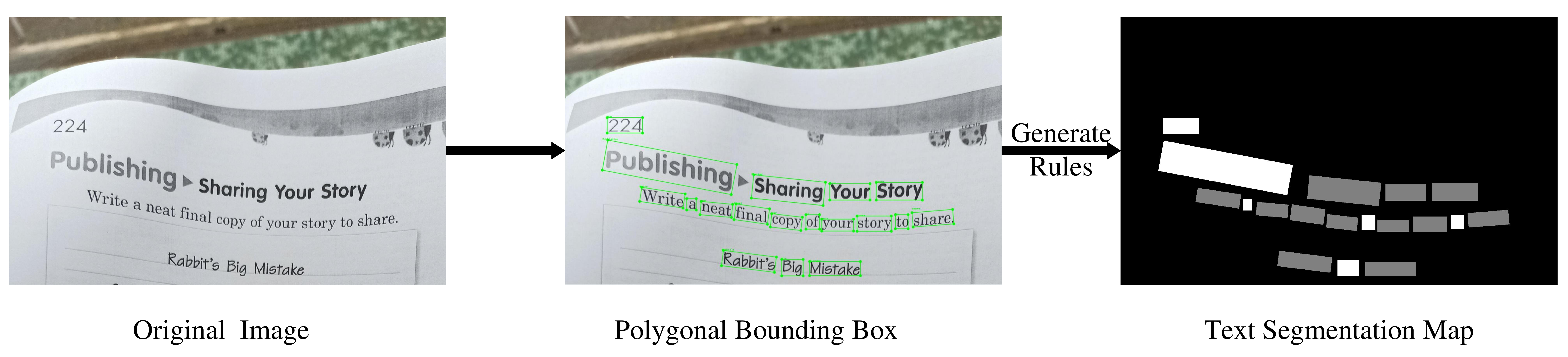}
		\caption{The generation of text segmentation map label for word Length-aware Segmentation Head. The generation rules are shown in Eq. (\ref{generation_rules}).}
		\label{text_attention_map_figure}
	\end{figure*}

	In dense text scene, we fully explore the information of text instances. As shown in Fig. \ref{text_attention_map_figure}, we differentiate text instances based on the number of characters and generate labels required for LenSeg Head based on existing ground truth.
	
	For a given image containing $Y$ text instances, each text instance is composed of a polygon, and the number of vertices of the polygon is determined by the characteristics of the dataset. The number of characters transcribed for each text instance $N_i$ determines the pixel value $v^{pix}_i$ of the area where the instance is located in the text segmentation map label $\mathbf{G}_{map}\in \mathbb{R}^{1\times W\times H}$. Text segmentation map label generation follows the rules in Sec. \ref{Section_3_3_2}:
	\begin{equation}
		v^{pix}_i=\begin{cases} 
			0.5,  & 4\leq N_i \leq 10 \\
			1, & 0 < N_i < 4 \ and \ 10 < N_i < N_{max},
		\end{cases}
		\label{generation_rules}
	\end{equation}
	where the initialization pixel value of the text segmentation map label $\mathbf{G}_{map}$ is all $0$, and the pixel value of the background area $v^{pix}_0$ is $0$. We define text instances that are less than $4$ characters and greater than $10$ characters as extreme word length text, with a pixel value of $1$ in the region. Other text instances are regular text, with a pixel value of $0.5$ in the area. For a given image containing $Y$ text instances, note that the text segmentation map label is a set of pixel values for the background and all text instances:  
	\begin{equation}
		\mathbf{G}_{map}=v^{pix}_0\cup v^{pix}_1\cup v^{pix}_2\cup \dots \cup v^{pix}_Y.
	\end{equation}
	
	\subsection{Optimization} \label{Section_3_5}
	
	During the detection phase, bipartite matching is used to obtain one-to-one matching between predictions and ground truth, and the loss is calculated as the weight of the matcher as follows:
	\begin{equation}
		\mathcal{L}_{det}=\lambda_{cls}\cdot \mathcal{L}_{cls}+\mathcal{L}_{boxes}+\mathcal{L}_{mask}+\lambda_{prior}\cdot \mathcal{L}_{prior}+\lambda_{map}\cdot \mathcal{L}_{map},
	\end{equation}
	where $\lambda_{cls}$, $\lambda_{prior}$, and $\lambda_{map}$ are hyper-parameters used to balance the loss. 
	
	The classification loss $\mathcal{L}_{cls}$ is a focal loss. More specifically, for the $k$-th text query, the probability of having a pure text instance class $\mathbf{\hat{p}}_k$. The focal loss used for text instance classification is defined as \cite{zhang2022text}:
	\begin{equation}
		\begin{aligned}
			\mathcal{L}_{cls}^k = & -\mathds{1}_{\{k\in \mathcal{I}m(\psi)\}}\alpha(1-\mathbf{\hat{p}}_k )^\gamma \log{(\mathbf{\hat{p}}_k )} \\
			& -\mathds{1}_{\{k\notin \mathcal{I}m(\psi)\}} (1-\alpha)(\mathbf{\hat{p}}_k )^\gamma \log{(\mathbf{1-\hat{p}}_k)},
		\end{aligned}
	\end{equation}
	where $\mathds{1}$ is the indicator function, $\mathcal{I}m(\psi)$ is the image of the mapping $\psi$. 
	
	Regression bounding boxes coordinate loss $\mathcal{L}_{boxes}$ uses $\ell_1$ loss and generalized IoU loss \cite{rezatofighi2019generalized}:
	\begin{equation}
		\mathcal{L}_{boxes}=\lambda_{\ell_1}\cdot \mathcal{L}_{\ell_1}+\lambda_{gIoU}\cdot \mathcal{L}_{gIoU}.
	\end{equation}
	
	Following \cite{huang2022swintextspotter}, the text mask loss $\mathcal{L}_{mask}$ uses $\ell_2$ loss and dice loss \cite{milletari2016v}:
	\begin{equation}
		\mathcal{L}_{mask}=\lambda_{\ell_2}\cdot \mathcal{L}_{\ell_2}+\lambda_{dice}\cdot \mathcal{L}_{dice}.
	\end{equation}
	
	The text word length information prediction task we introduced calculates the loss $\mathcal{L}_{prior}$ using Smooth $\ell_1$ loss, and the calculation formula is as follows:
	\begin{equation}
		\mathcal{L}_{prior}=\text{Smooth}_{\ell_1} (\mathbf{G}_{prior}-\mathbf{C}_{prior} ),
	\end{equation}
	where $\mathbf{C}_{prior}$ is the set of predicted values for the text word length information. 
	
	On the other hand, the word length-aware segmentation loss $\mathcal{L}_{map}$ we introduced also uses dice loss, and the calculation formula is as follows:
	\begin{equation}
		\mathcal{L}_{map}=1-\frac{2|\mathbf{G}_{map}\cap \mathbf{C}_{map}|}{|\mathbf{G}_{map}|+|\mathbf{C}_{map}|},
	\end{equation}
	where $\mathbf{C}_{map}$ is the predicted text segmentation map, $|\mathbf{G}_{map}\cap \mathbf{C}_{map}|$ is the intersection between $\mathbf{G}_{map}$ and $\mathbf{C}_{map}$, and $|\mathbf{G}_{map}|$ and $|\mathbf{C}_{map}|$ represent the number of elements of  $\mathbf{G}_{map}$ and $\mathbf{C}_{map}$.
	
	During the recognition phase, the loss $\mathcal{L}_{rec}$ is described as:
	\begin{equation}
		\mathcal{L}_{rec}=-\frac{1}{T}\sum_{j=1}^T \log{p(\mathbf{y}_j)},
	\end{equation}
	where $T$ is the length of the sequence label, and $p(\mathbf{y}_j)$ is the probability of the sequence \cite{liao2021mask}. 
	
	
	\subsection{WordLenSpotter training pipeline} \label{Section_3_6}
	
	We summarize the WordLenSpotter training pipeline in Alg. \ref{alg_1}. Firstly, we use the image encoder to extract image features $\mathbf{f}_{im}$ and initialize the text queries $\mathbf{q}^0$ along with the text bounding box coordinates $\mathbf{b}^0$. In each $i$-th ($i = 1,\cdots, N$) detection stage, we iteratively refine the RoI features $\mathbf{f}_{RoI}^i$ based on the previous predicted bounding boxes. We then update the text queries $\mathbf{q}^i$ using the Detection Decoder, and also update the bounding boxes $\mathbf{b}^i$, instance class $\mathbf{c}^i$, and text mask $\mathbf{m}^i$ through the Detection Head, update the text word length prediction information $\mathbf{C}_{prior}$ via SLP, update the text segmentation map $\mathbf{C}_{map}$ with LenSeg Head. Using  $\mathbf{b}^i$, $\mathbf{c}^i$, $\mathbf{m}^i$, $\mathbf{C}_{prior}$ and $\mathbf{C}_{map}$ to match the ground truth, we obtain the detection loss $\mathcal{L}_{det}^i$, which is employed to update the detection part of the network parameters $\boldsymbol{\theta}^w_{det}$ using the AdamW optimizer. 
	
	\begin{algorithm}[t]
		\caption{WordLenSpotter Training}
		\label{alg_1}
		\SetKwData{Image}{Image} \SetKwData{GT}{GT}
		\SetKwData{Learning}{Learning rate $\boldsymbol{\eta}$} \SetKwData{$f_{im}$}{$\mathbf{f}_{im}$}
		\SetKwData{$f_{rec}$}{$f_{rec}$} \SetKwData{$f_{RoI}^i$}{$f_{RoI}^i$} 
		\SetKwData{$f_{RoI}^N$}{$f_{RoI}^N$} 
		\SetKwData{$b^0$}{$b^0$}
		\SetKwData{$q^0$}{$q^0$}
		\SetKwData{$b^i$}{$b^i$}
		\SetKwData{$q^i$}{$q^i$}
		\SetKwData{$q^N$}{$q^N$}
		\SetKwData{$q^{N+1}$}{$q^{N+1}$}
		\SetKwData{$p_{wl}^i$}{$p_{wl}^i$}
		\SetKwData{$q^i$}{$q^i$}
		\SetKwData{$M_t^i$}{$M_t^i$}
		\SetKwData{$M_t^N$}{$M_t^N$}
		\SetKwData{$C_t^i$}{$C_t^i$}
		\SetKwData{$L$}{$\mathcal{L}$}
		\SetKwData{$L_{det}^i$}{$\mathcal{L}_{det}^i$}
		\SetKwData{$L_{rec}$}{$\mathcal{L}_{rec}$}
		\SetKwData{$P_{rec}$}{$P_{rec}$}
		\SetKwFunction{ImageEncoder}{ImageEncoder} \SetKwFunction{RoIAlign}{RoIAlign} \SetKwFunction{DetectionDecoder}{DetectionDecoder}
		\SetKwFunction{SLP}{SLP}
		\SetKwFunction{LenSegHead}{LenSegHead}
		\SetKwFunction{DetectionHead}{DetectionHead}
		\SetKwFunction{HungarianMatcher}{HungarianMatcher}
		\SetKwFunction{FeatureFusion}{FeatureFusion}
		\SetKwFunction{RecognitionHead}{RecognitionHead}
		\SetKwFunction{LossSeqDecoder}{LossSeqDecoder}
		\KwIn{\Image, \GT, \Learning}
		\KwOut{Optimized network parameters \ $\boldsymbol{\theta}^w_{det}$,  $\boldsymbol{\theta}^w$}
		\emph{Initialize \ $\mathbf{q}^0$, \ $\mathbf{b}^0$ randomly}\;
		$\mathbf{f}_{im}$ $\leftarrow$ \ImageEncoder (\Image)\;
		\For{$i = 1$ \KwTo $N$}{
			\ $\mathbf{f}_{RoI}^i$ $\leftarrow$ \RoIAlign{\ $\mathbf{f}_{im}$, \ $\mathbf{b}^{i-1}$}\;
			\emph{Text queries update} \ $\mathbf{q}^i$ $\leftarrow$ \DetectionDecoder{\ $\mathbf{f}_{RoI}^i$, \ $\mathbf{q}^{i-1}$}\;
			\ $\mathbf{b}^i$, \ $\mathbf{c}^i$, \ $\mathbf{m}^i$ $\leftarrow$ \DetectionHead{$\mathbf{q}^i$}\;
			\ $\mathbf{C}_{prior}$ $\leftarrow$ \SLP{\ $\mathbf{f}_{RoI}^i$}\;
			\ $\mathbf{C}_{map}$ $\leftarrow$ \LenSegHead{\ $\mathbf{q}^i$}\;
			\ $\mathcal{L}_{det}^i$ $\leftarrow$ \HungarianMatcher{\{ $\mathbf{b}^i$, \ $\mathbf{c}^i$, \ $\mathbf{m}^i$, \ $\mathbf{C}_{prior}$, \ $\mathbf{C}_{map}$\}, \GT}\;
			\emph{Update detection parameters} $\boldsymbol{\theta}^w_{det}$ $\leftarrow$ $\boldsymbol{\theta}^w_{det} - \boldsymbol{\eta} \nabla_{\boldsymbol{\theta}^w_{det}} \cdot \mathcal{L}^i_{det}$ \;}
		$\mathbf{f}_{RoI}^{N+1}$ $\leftarrow$ \RoIAlign{\ $\mathbf{f}_{im}$, \ $\mathbf{b}^N$}\;
		\emph{Text queries update} $\mathbf{q}^{N+1}$ $\leftarrow$ \DetectionDecoder{\ $\mathbf{f}_{RoI}^{N+1}$, \ $\mathbf{q}^N$}\;
		$\mathbf{f}_{rec}$ $\leftarrow$ \FeatureFusion{\ $\mathbf{f}_{RoI}^{N+1}$, \ $\mathbf{q}^{N+1}$}\;
		$\mathcal{L}_{rec}$ $\leftarrow$ \RecognitionHead{\{ $\mathbf{f}_{rec}$, $\mathbf{m}^N$\}, \GT}\;
		\emph{Update overall model parameters} $\boldsymbol{\theta}^w$ $\leftarrow$ $\boldsymbol{\theta}^w - \boldsymbol{\eta} \nabla_{\boldsymbol{\theta}^w} \cdot \mathcal{L}_{rec}$ \;
	\end{algorithm}
	
	In the $(N+1)$-th stage, we acquire refined RoI features $\mathbf{f}_{RoI}^{N+1}$, and text queries $\mathbf{q}^{N+1}$ based on the predicted bounding boxes $\mathbf{b}^N$ from the $N$-th stage, and fuses them at multiple scales to produce recognition features $\mathbf{f}_{rec}$. Finally, we use the Recognition Head to supervise the recognition features $\mathbf{f}_{rec}$, the text mask $\mathbf{m}^N$ predicted in the $N$-th detection stage. We calculate the recognition loss $\mathcal{L}_{rec}$, which is then used to update the overall model parameters $\boldsymbol{\theta}^w$ via AdamW optimizer.
	
	In summary, our approach iteratively optimizes the detection parameters $\boldsymbol{\theta}^w_{det}$ for the previous $N$ stages using the detection loss $\mathcal{L}_{det}^i$. Additionally, we optimize the overall model parameters $\boldsymbol{\theta}^w$ in the $N+1$ stage using the recognition loss $\mathcal{L}_{rec}$. This iterative process enables us to achieve a well-performing network for WordLenSpotter.
	
	\section{Experiments} \label{Section_4}
	
	\begin{figure}[t]
		\centering
		\includegraphics[width=\linewidth]{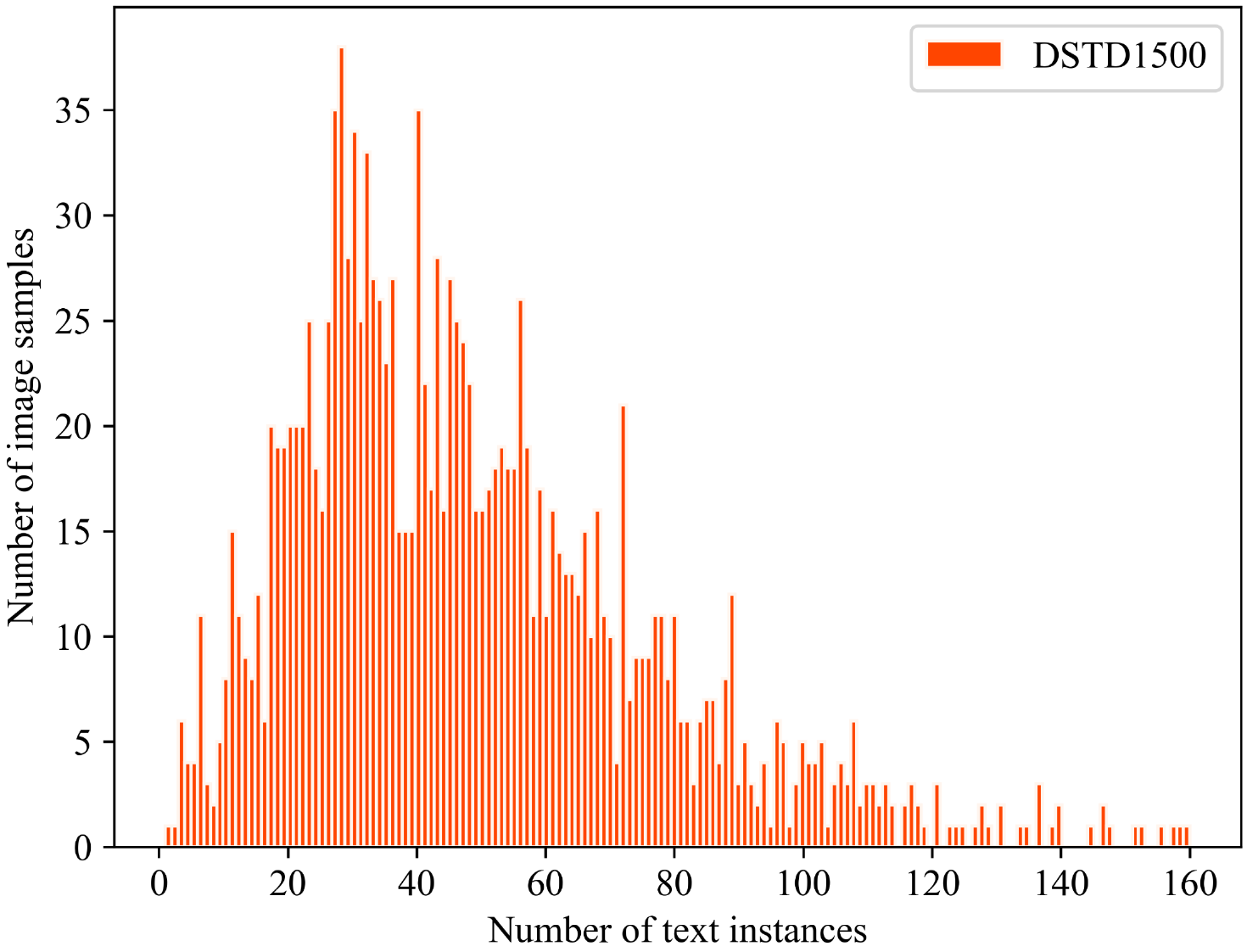}
		\caption{Figure displaying the analysis of DSTD$1500$ data samples. A higher bar indicates that there are more images with that number of instances.}
		\label{dstd1500_data_analysis_figure}
	\end{figure}
	
	\subsection{Datasets} \label{Section_4_1}
	
	\begin{figure*}[tbp]
		\centering 
		\includegraphics[width=\linewidth]{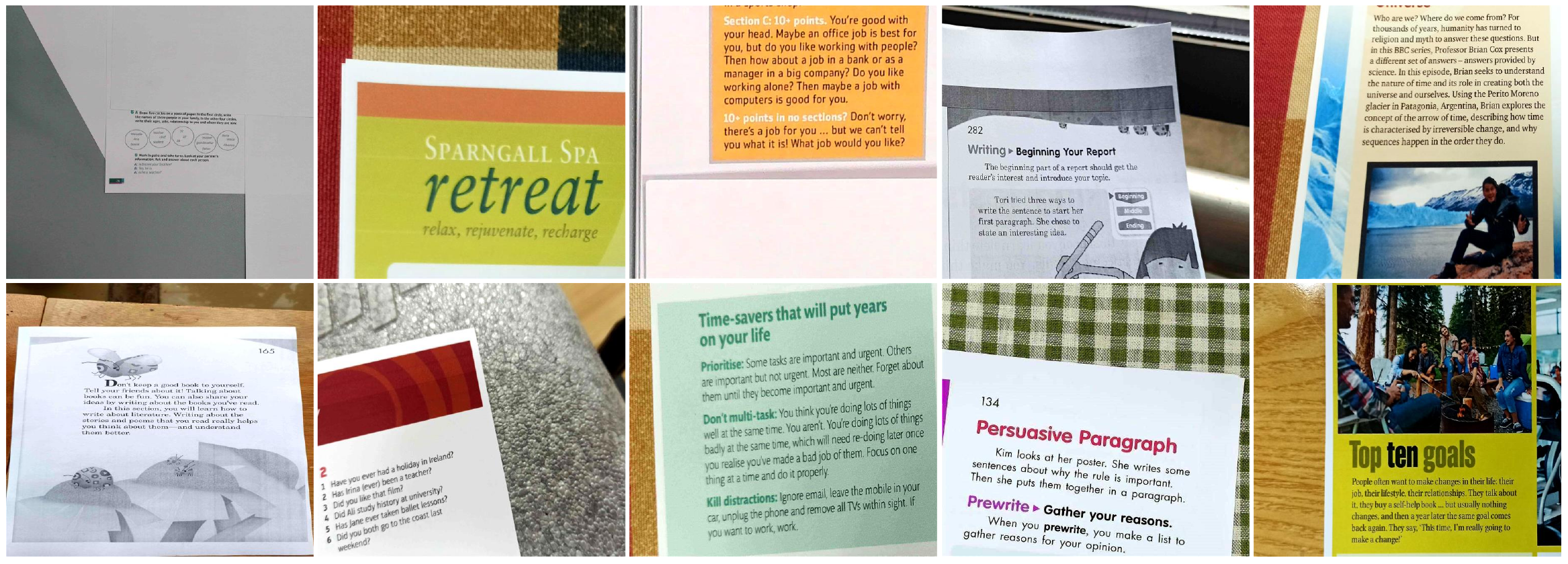}
		\caption{DSTD$1500$ partial sample example. The text within the image is densely distributed and encompasses text samples of various scales.}
		\label{dataset_fig_small_figure}
	\end{figure*}
	
	\subsubsection{Benchmark} \label{Section_4_1_1}
	
	\textbf{SynthText $\mathbf{150}$k} \cite{liu2020abcnet} includes $94,723$ multi-oriented straight text images and $54,327$ curved text images, with word level bounding boxes and annotation. \textbf{ICDAR $\mathbf{2013}$} \cite{pratikakis2013icdar} is a scene text dataset consisting of $229$ training images and $233$ test images. \textbf{ICDAR $\mathbf{2015}$} \cite{karatzas2015icdar} contains $1,000$ training images and $500$ test images, with multi-oriented text annotated with quadrilateral boxes. \textbf{Total-Text} \cite{ch2020total} is a benchmark for arbitrarily shaped scene text, using word level polygonal boxes as annotations, with $1,255$ images for training and $300$ images for testing. \textbf{CTW$\mathbf{1500}$} \cite{liu2019curved} is a scene text benchmark of any shape at the text line level, consisting of $1,000$ training images and $500$ test images, including $3,530$ curved and multi-oriented text instances. \textbf{ICDAR $\mathbf{2017}$ MLT} \cite{nayef2017icdar2017} is a multi-oriented multilingual scene text dataset. It contains $7,200$ training images and $1,800$ validation images. We only select images containing English text for training.
	
	\subsubsection{Dense scene text dataset} \label{Section_4_1_2}
	
	To evaluate the performance of the proposed WordLenSpotter network in text spotting in dense scenes, we contribute a dense scene text dataset called DSTD$1500$ from real reading scenes. The dataset consists of $1500$ images captured from paper books, along with $75,471$ English text instances annotated with polygonal bounding boxes and corresponding transcriptions of text characters. Fig. \ref{dstd1500_data_analysis_figure} shows the density distribution of text instances in each image of the DSTD$1500$. The majority of images contain $30$ or more text instances. In Fig. \ref{dataset_fig_small_figure}, we show several sample images. These images contain densely distributed and diverse text instances.
	
	\subsection{Implementation details} \label{Section_4_2}
	
	Our WordLenSpotter network is implemented in Python and utilizes four NVIDIA Tesla A$800$ GPUs for efficient processing. The number of text queries we typically use $Q_{num}=600$. The maximum text length $N_{max}=25$, for CTW$1500$, in order to solve the problem of longer text in the dataset, the maximum text length $N_{max}$ is set to $100$. Our model predicts $96$ character categories. Loss weight factor $\lambda_{cls}=2.0$, $\lambda_{prior}=1.0$, $\lambda_{map}=2.0$, $\lambda_{\ell_1}= 5.0$, $\lambda_{gIoU}=2.0$, $\lambda_{\ell_2}=\lambda_{dice}=2.0$, the parameter setting for the focal loss is $\alpha=0.25$, $\gamma=2.0$. We set the batch size to $8$ and utilize the AdamW optimizer to train our network end-to-end.
	
	The structural details of the network model are as follows: 
	
	The image encoder employs an embedding dimension of $96$, and a window size of $7$ SwinTransformer. The attention block consists of $4$ layers, and the number of heads for multi-head attention is $\{3, 6, 12, 24\}$, respectively. In addition, $3\times3$ depthwise convolutions and $1\times1$ pointwise convolutions as the convolutional layer. The dilated convolution module uses two dilated convolutions with dilation rates of $2$ and $4$, respectively. The output channels are all $256$. Upsampling in this context uses the nearest neighbor interpolation method with a scaling factor set at $2$, while downsampling is achieved through convolution with a stride of $2$. Between every two layers, use batch normalization and ReLU processing. 
	
	The detection decoder uses a query based multi head attention layer with $6$ layers of $8$ heads and $256$ hidden dimensions, and the dropout rate is $0.1$. The dynamic convolutional layer follows the configuration of SparseR-CNN \cite{sun2021sparse}, with a hidden dimension of 256, a dynamic dimension of 64, and a number of dynamic heads of 2. 
	
	SPL incorporates three layers comprising $3\times3$ standard convolutions, batch normalization, and ReLU, resulting in an output channel of $256$. Subsequently, a linear layer composed of linear mapping, layer normalization, and ELU is employed to predict word length information, with an output channel of $2$. 
	
	LenSeg Head without any convolution operations. It sequentially employs two layers of linear mapping, layer normalization, and ELU, resulting in an output scale of $224\times224$. In addition, multiscale feature fusion and recognition heads follow the settings of SwinTextSpotter \cite{huang2022swintextspotter}.
	
	\begin{table}[t]
		\scriptsize
		\centering
		\caption{Dense scene text spotting results on DSTD$1500$. “None” represents lexicon-free, while “Full” indicates all the words in the test set are used.}
		\label{dstd1500_table}
		\begin{tabular}{lcccccc}
			\toprule
			\multirow{2}{*}{Method}  & \multicolumn{3}{c}{Detection} & \multicolumn{2}{c}{End-to-End} \\ \cmidrule{2-6} 
			& P        & R        & F       & None           & Full          \\ \midrule
			PAN++ \cite{wang2021pan++}          & -     & -     & -     & $68.00$ & -     \\
			ABCNetV$2$ \cite{liu2021abcnet}         & $81.77$ & $70.12$ & $75.50$ & $70.86$ & $73.40$ \\
			TESTR \cite{zhang2022text}           & $82.33$ & $75.10$ & $78.55$ & $71.55$ & $73.48$ \\
			SPTS \cite{peng2022spts}            & -     & -     & -     & $72.28$ & $72.82$ \\
			SwinTextSpotter \cite{huang2022swintextspotter} & $86.01$ & $78.00$ & $81.81$ & $74.57$ & $81.45$ \\
			\textbf{WordLenSpotter }                               & $\mathbf{88.64}$ & $\mathbf{78.57}$ & $\mathbf{83.30}$ & $\mathbf{75.16}$ & $\mathbf{81.95}$ \\ \bottomrule
		\end{tabular}
	\end{table}
	
	In addition, we also adopt the following data augmentation strategy: ($1$) Random scaling, specifying the unique selection of short sizes from $640$ to $896$ (with an interval of $32$), and long sizes not exceeding $1,600$; ($2$) Random cropping ensures that the cropped image does not cut text instances; ($3$) Random rotation, we rotate the image between positive and negative $90$ degrees. During the training period, other data augmentation strategies are also applied, such as random brightness, contrast, and saturation.
	
	Firstly, we pre-train WordLenSpotter with $450$K iterations on the entire SynthText $150$k, ICDAR $2017$ MLT, and corresponding benchmark training sets. The initial learning rate is $2.5\times10^{-5}$, drop to $2.5\times10^{-6}$ in the $360$K iteration, and drop to $2.5\times10^{-7}$ at $420$K iteration. Then we jointly train the model on ICDAR $2013$, ICDAR $2015$, ICDAR $2017$ MLT, Total Text, and DSTD$1500$ for $150$K iterations, and attenuate to one-tenth at the $100$K iteration. Finally, we fine-tune the jointly trained model on the corresponding dataset.
	
	\subsection{Comparison with state-of-the-art} \label{Section_4_3}
	
	\subsubsection{Results on DSTD1500} \label{Section_4_3_1}
	
	To evaluate the performance of WordLenSpotter in text detection and end-to-end spotting in dense scenes, we compare its quantitative results with state-of-the-art methods. The results presented in Table \ref{dstd1500_table} are obtained by fine-tuning on DSTD$1500$ using the original training program. We use standard recall rate, precision rate, and F-score as evaluation metrics.
	
	\begin{figure*}[t]
		\centering 
		\includegraphics[width=\linewidth]{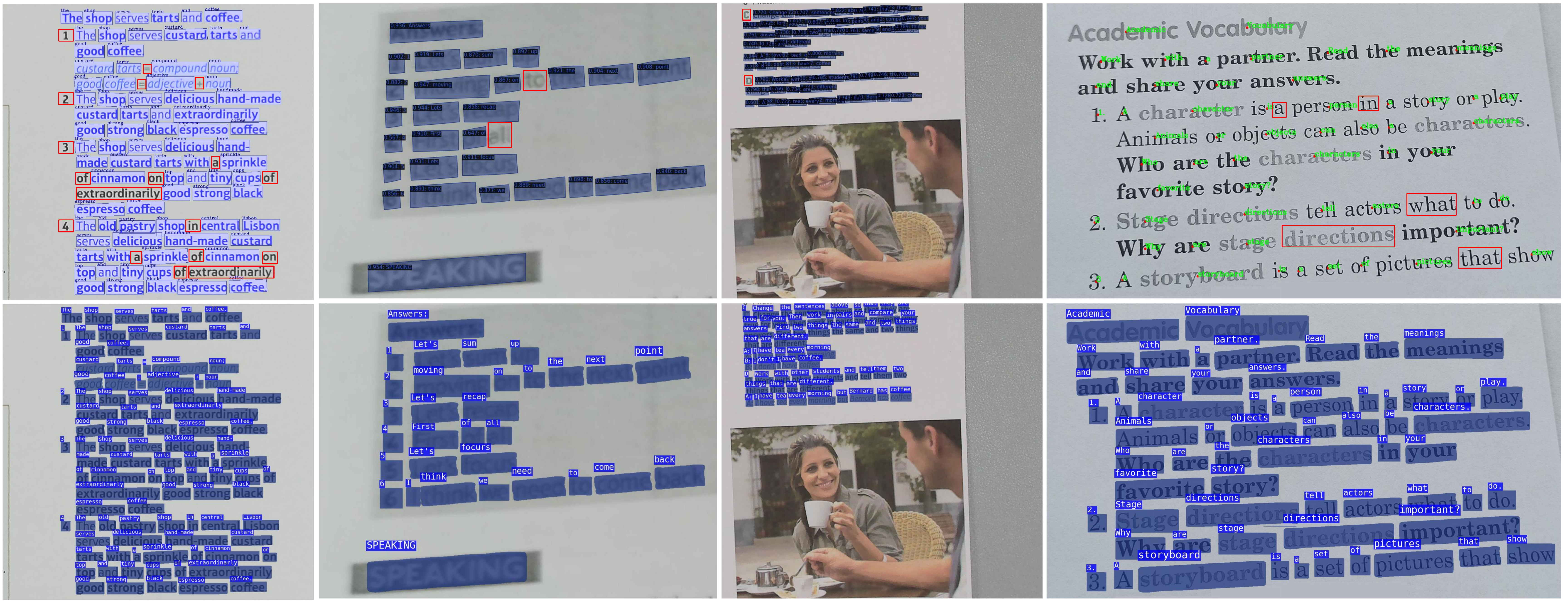}
		\caption{The qualitative results of the state-of-the-art methods (the first row is PAN++ \cite{wang2021pan++}, TESTR \cite{zhang2022text}, ABCNetv$2$ \cite{liu2021abcnet} and SPTS \cite{peng2022spts} from left to right.) and WordLenSpotter (the second row) on DSTD$1500$ without the lexicon. Red rectangular boxes indicate missed detections.}
		\label{dstd1500_vis_figure}
	\end{figure*}
	
	\begin{figure*}[t]
		\centering 
		\begin{minipage}[t]{0.3\linewidth}
			\centering
			\includegraphics[width=0.95\textwidth]{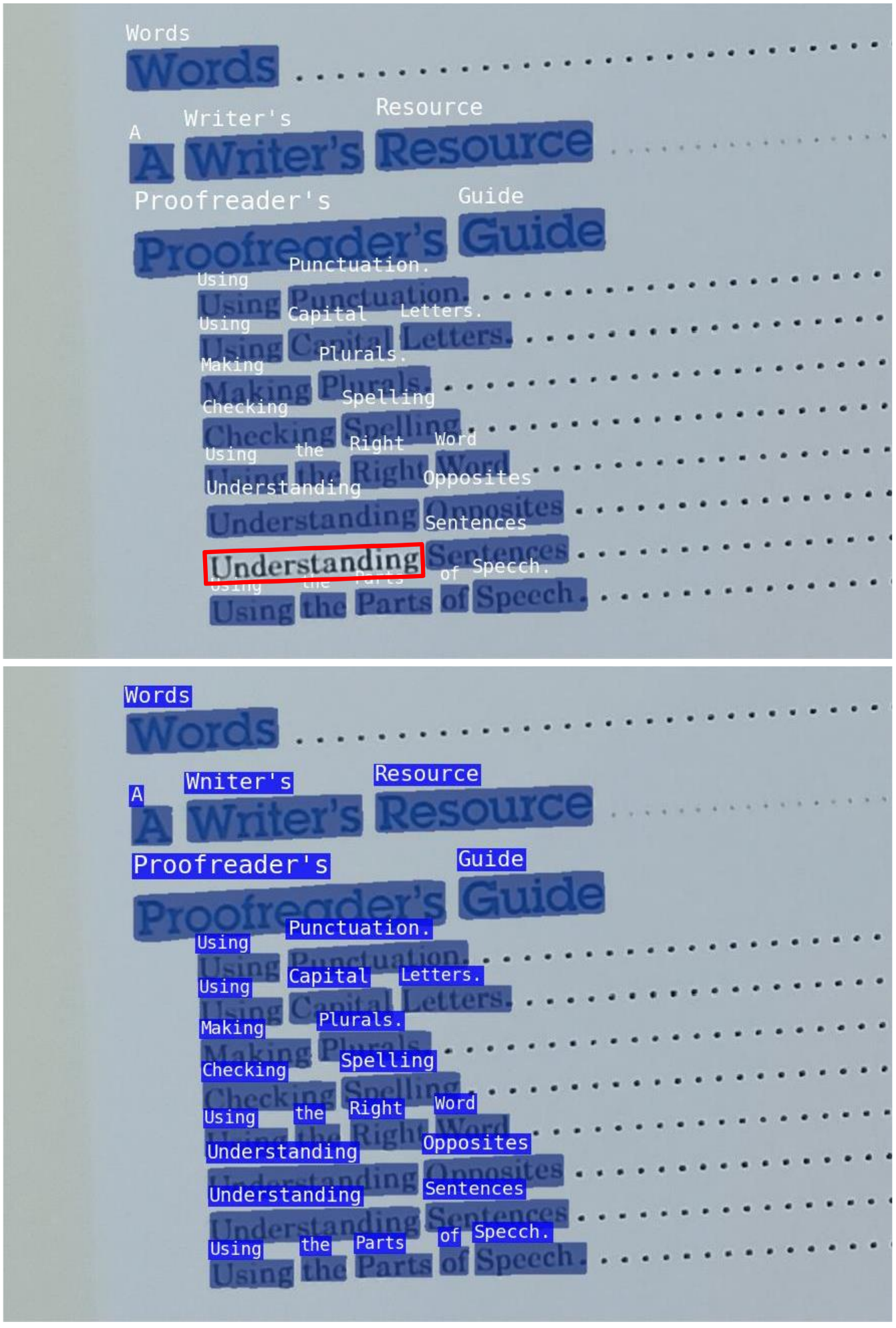}
			\centerline{(a) Long text sample.}
		\end{minipage}
		\begin{minipage}[t]{0.3\linewidth}
			\centering
			\includegraphics[width=0.98\textwidth]{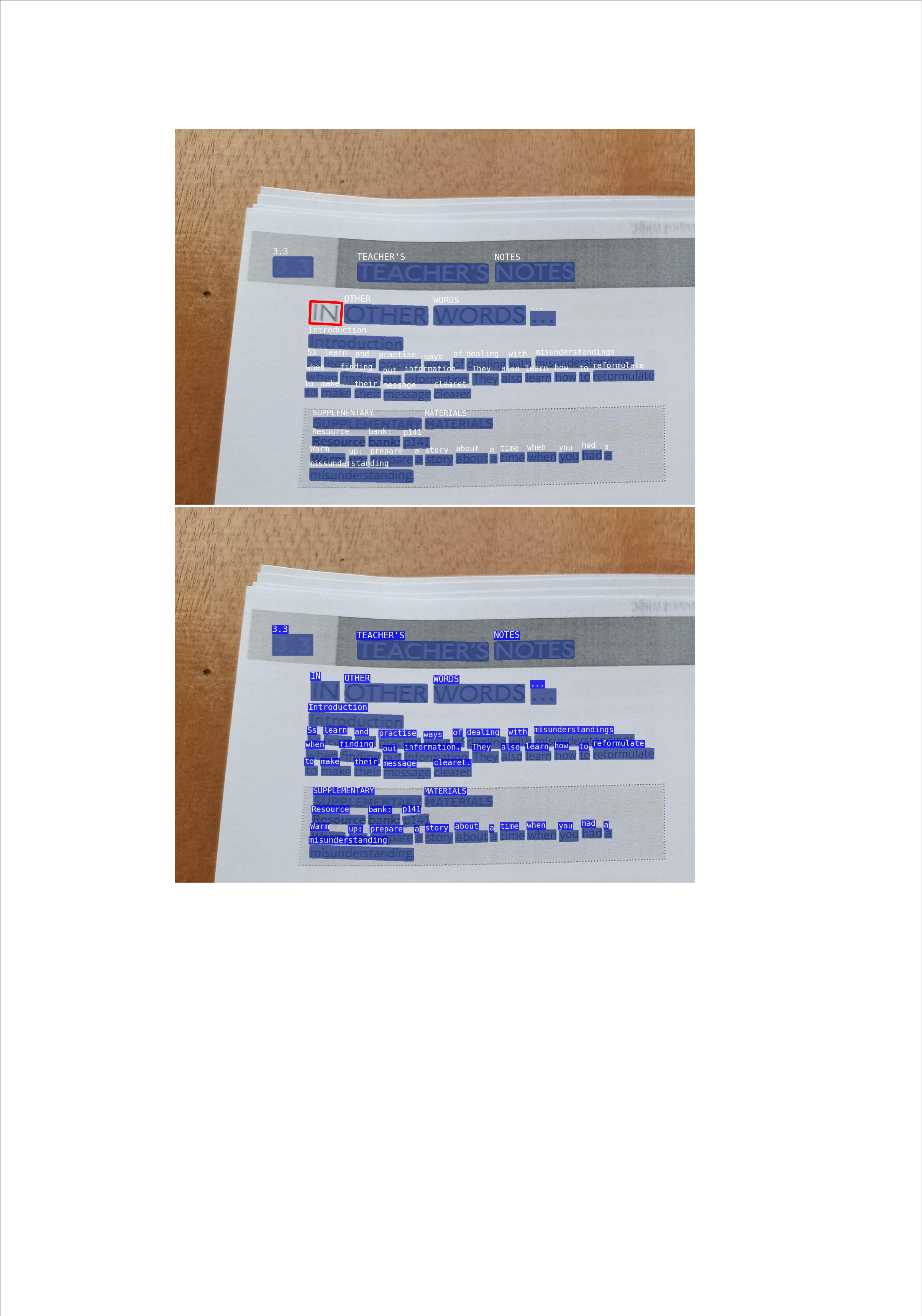}
			\centerline{(b) Short text sample.}
		\end{minipage}
		\begin{minipage}[t]{0.01\linewidth}
			\centering
		\end{minipage}
		\begin{minipage}[t]{0.32\linewidth}
			\centering
			\includegraphics[width=0.98\textwidth]{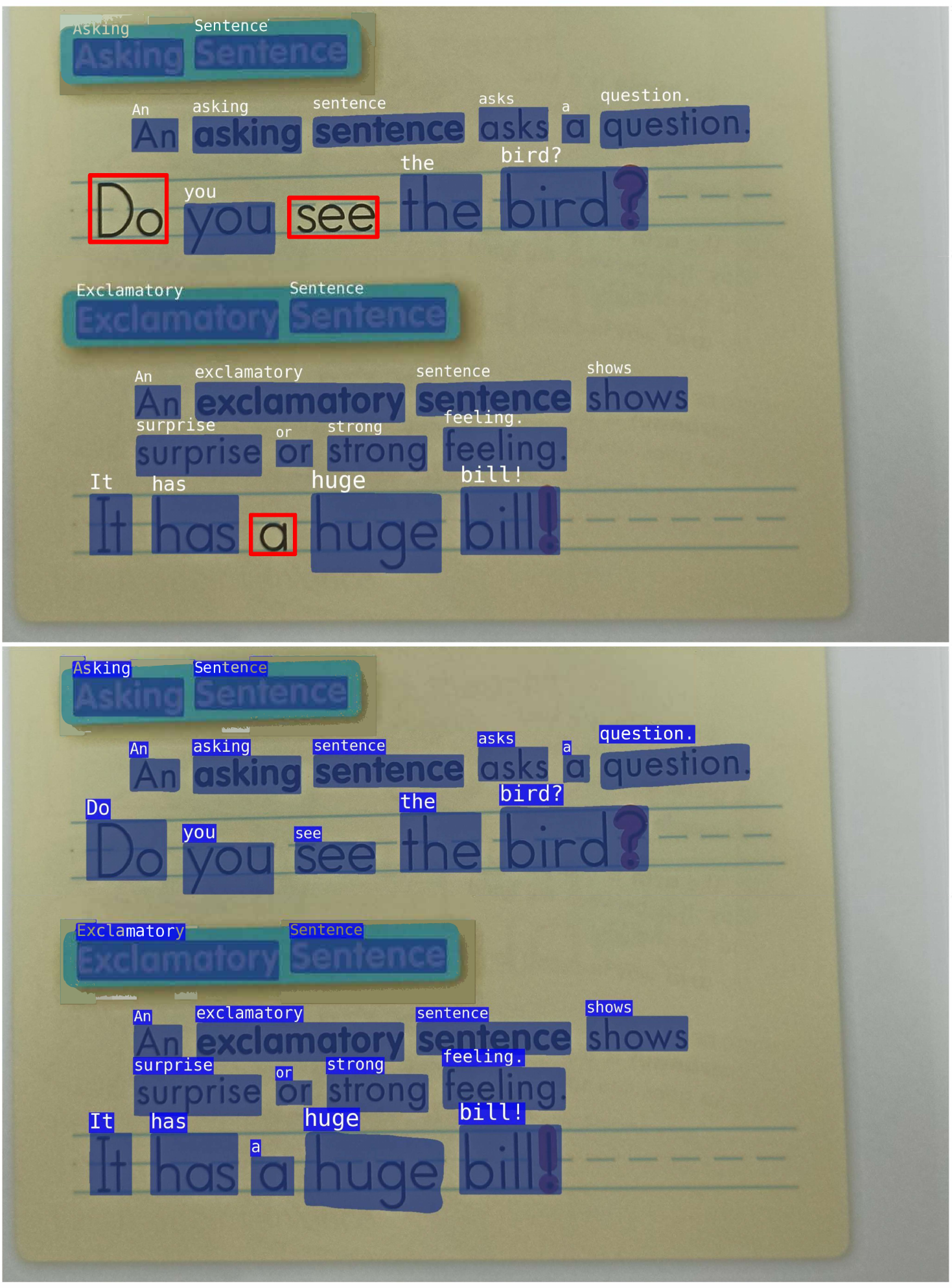}
			\centerline{(c) Multiscale text sample.}
		\end{minipage}
		\caption{Comparison of qualitative results between baseline \cite{huang2022swintextspotter} (top) and WordLenSpotter (bottom) on DSTD1500 without the lexicon.}
		\label{dstd1500_vis_01_figure}
	\end{figure*}
	
	In terms of text detection, WordLenSpotter achieves an F-score of $83.30\%$ on the dense scene text dataset DSTD$1500$, which is $1.49\%$ higher than existing advanced methods, while demonstrating higher precision and recall. In the case of end-to-end text spotting, WordLenSpotter also performs best when the DSTD$1500$ is equipped with a complete lexicon, with an F-score of $81.95\%$, which is at least $0.50\%$ higher than other methods. When DSTD$1500$ does not use lexicon, the F-score of WordLenSpotter is at least $0.59\%$ higher than other methods, reaching $75.16\%$.
	
	\begin{figure}[t]
		\centering 
		\includegraphics[width=\linewidth]{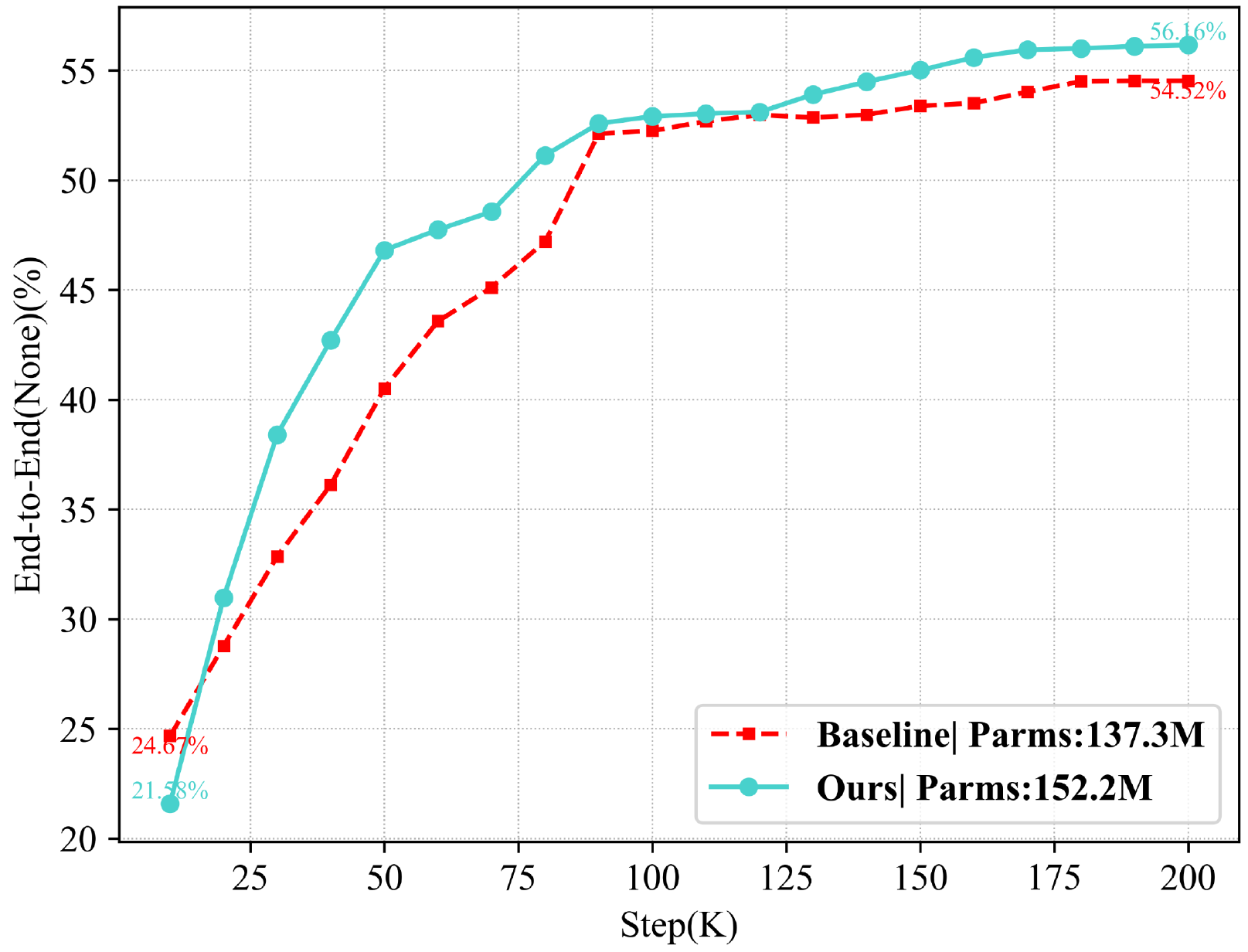}
		\caption{Comparison between WordLenSpotter and baseline \cite{huang2022swintextspotter} on only the DSTD$1500$ training set. Start training from scratch for $200$K iterations respectively, and record the validation results in every $10$K iterations. Parms represents the parameter quantity of the network model.}
		\label{Compare_plot_figure}
	\end{figure}
	
	Fig. \ref{dstd1500_vis_figure} shows the qualitative results of the state-of-the-art methods and WordLenSpotter in DSTD$1500$. The results show that our proposed WordLenSpotter can effectively handle densely distributed and numerous text images in dense scenes, with a high recall rate for extreme word length text instances, and reduce the missed detection phenomenon of text spotting in dense scenes. In short, the experimental results on DSTD$1500$ demonstrate the effectiveness of our method in text localization in dense scenes.
	
	In order to demonstrate the effectiveness of our proposed text word length information prediction branch and specialized word length-based segmentation head, we compare the qualitative results of baseline \cite{huang2022swintextspotter} and WordLenSpotter on DSTD$1500$ for spotting short and long text instances in dense scenes with long-tailed distributions. The results indicate that our method can better spot text instances with extreme word length in dense scenes, such as Fig. \ref{dstd1500_vis_01_figure}(a) and (b). Moreover, our method demonstrates improved detection of text instances across different scales, as illustrated in Fig. \ref{dstd1500_vis_01_figure}(c). This further validates the effectiveness of our image encoder.
	
	We conduct a comparative experiment, evaluating the baseline \cite{huang2022swintextspotter} and WordLenSpotter separately, using $200$K iterations on the DSTD$1500$ training set. Unless stated otherwise, the number of queries in our experiment is set to $600$. The experimental results include network parameter quantities for the baseline and our proposed method, as well as end-to-end spotting results without the lexicon on DSTD$1500$. In Fig. \ref{Compare_plot_figure}, our method exhibits rapid convergence and better performance within a reasonable range of network parameters ($56.16\%$ vs. $54.52\%$). 
	
\begin{figure*}[tp]
		\centering 
		\includegraphics[width=0.92\linewidth]{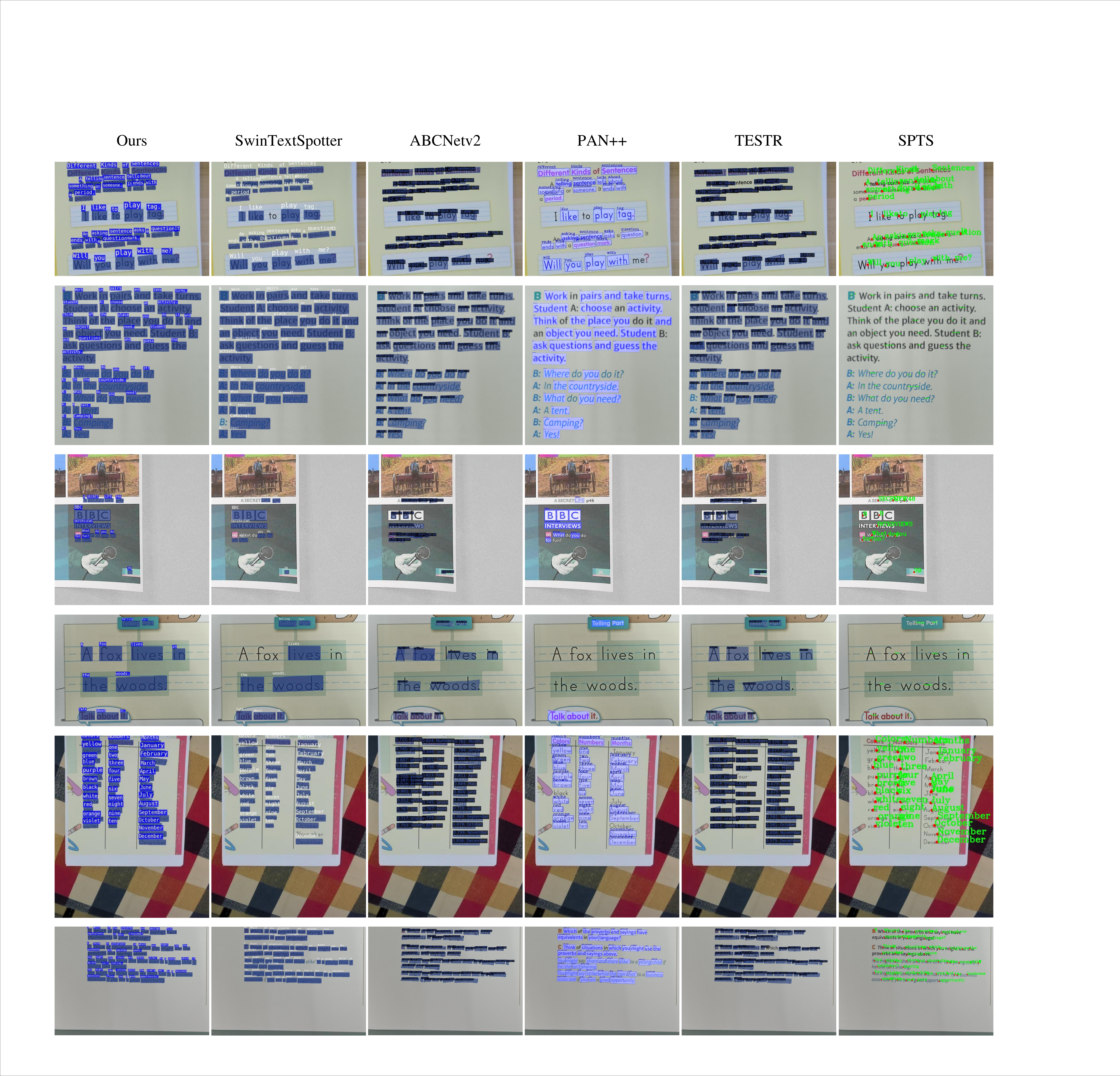}
		\caption{Comparison of visualization results on DSTD$1500$. Ours: WordLenSpotter. SwinTextSpotter: our baseline \cite{huang2022swintextspotter}, in order to have better performance in dense text images, we propose setting the number of proposals to $600$. ABCNetv$2$: we train and fine-tune ABCNetv$2$ \cite{liu2021abcnet} after generating Bezier annotations for DSTD$1500$. PAN++: for the sake of fairness, we retain text instances that are successfully detected and recognized through the PAN++ \cite{wang2021pan++}. TESTR: we generate polygon annotations containing $16$ control points for the DSTD$1500$ to train and fine-tune the TEST-Polygon \cite{zhang2022text}. SPTS: following the description of SPTS \cite{peng2022spts}, we utilize a single center point prediction method for training on the DSTD$1500$ dataset.}	\label{Compare_visualization_figure}
	\end{figure*}

	To ensure compatibility and expansion of WordLenSpotter with existing state-of-the-art methods, we generate training annotations suitable for model training on DSTD$1500$, following the different training annotation requirements of these methods. More visualization results on dense text images are shown in Fig. \ref{Compare_visualization_figure}.
	
	\begin{figure*}[t]
		\centering 
		\includegraphics[scale=0.15]{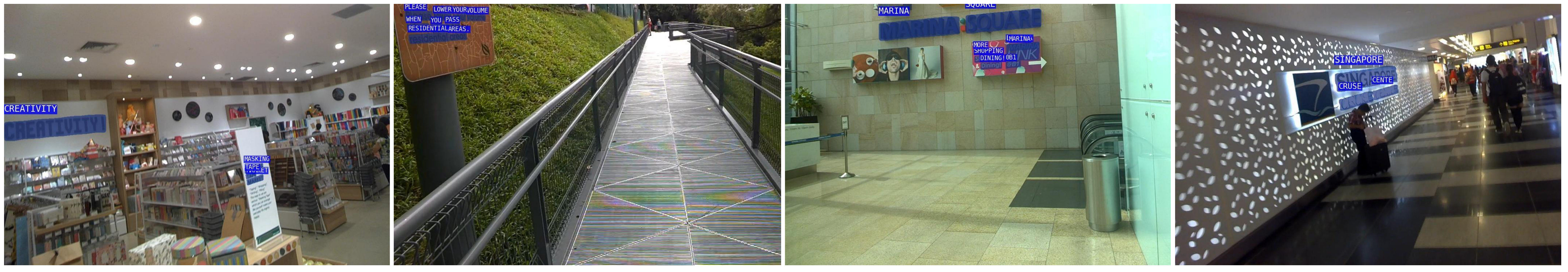}
		\caption{Qualitative results of WordLenSpotter on ICDAR$2015$.}
		\label{icdar2015_vis_figure}
	\end{figure*}
	
	\subsubsection{Results on benchmark} \label{Section_4_3_2}
	
	We evaluate WordLenSpotter on various popular benchmarks to demonstrate the competitiveness of our method in text detection and end-to-end spotting across different scenes.
	
	\begin{table}[t]
		\scriptsize
		\centering
		\caption{Scene text spotting results on ICDAR$2015$. “S”, “W”, and “G” represent recognition with “Strong”, “Weak”, and “Generic” lexicon, respectively.}
		\label{icdar2015_table}
		\begin{tabular}{lcccccc}
			\toprule
			\multirow{2}{*}{Method}  & \multicolumn{3}{c}{Detection} & \multicolumn{3}{c}{End-to-End} \\ \cmidrule{2-7} 
			& P        & R        & F       & S           & W      & G     \\ \midrule
			FOTS \cite{liu2018fots}              & $91.0$ & $85.2$ & $88.0$ & $81.1$ & $75.9$ & $60.8$ \\
			TextDragon \cite{feng2019textdragon}    & $\mathbf{92.5}$ & $83.8$ & $87.9$ & $82.5$ & $78.3$ & $65.2$ \\
			TextPerceptron \cite{qiao2020text}   & $92.3$ & $82.5$ & $87.1$ & $80.5$ & $76.6$ & $65.1$ \\
			Mask TextSpotter v$3$ \cite{liao2020mask}    & -    & -    & -    & $83.3$ & $78.1$ & $74.2$ \\
			PGNet \cite{wang2021pgnet}            & $91.8$ & $84.8$ & $88.2$ & $83.3$ & $78.3$ & $63.5$ \\
			MANGO \cite{qiao2021mango}           & -    & -    & -    & $85.4$ & $80.1$ & $73.9$ \\
			ABCNetV2 \cite{liu2021abcnet}         & $90.4$ & $86.0$ & $88.1$ & $82.7$ & $78.5$ & $73.0$ \\
			PAN++  \cite{wang2021pan++}         & -    & -    & -    & $82.7$ & $78.2$ & $69.2$ \\
			TESTR \cite{zhang2022text}          & $90.3$ & $\mathbf{89.7}$ & $\mathbf{90.0}$ & $85.2$ & $79.4$ & $73.6$ \\
			GLASS \cite{ronen2022glass}          & $86.9$ & $84.5$ & $85.7$ & $84.7$ & $80.1$ & $\mathbf{76.3}$ \\
			SwinTextSpotter \cite{huang2022swintextspotter} & -    & -    & -    & $83.9$ & $77.3$ & $70.5$ \\
			TTS \cite{kittenplon2022towards}       & -    & -    & -    & $85.2$ & $81.7$ & $77.4$ \\
			SPTS \cite{peng2022spts}            & -    & -    & -    & $77.5$ & $70.2$ & $65.8$ \\
			\textbf{WordLenSpotter}                                & $90.4$ & $87.9$ & $89.1$ & $\mathbf{85.5}$ & $\mathbf{82.1}$ & $75.1$ \\ \bottomrule
		\end{tabular}
	\end{table}
	
	We assess the performance of WordLenSpotter on multi-oriented text spotting on the multi-oriented text benchmark ICDAR$2015$. The quantitative results are shown in Table \ref{icdar2015_table}, and our method demonstrates a highly competitive performance when compared to other methods in the evaluation. Specifically, WordLenSpotter achieves $85.5\%$, $82.1\%$, and $75.1\%$ of the “S”, “W”, and “G” metrics in end-to-end text spotting tasks, and achieves state-of-the-art performance on “S” and “W”. Fig. \ref{icdar2015_vis_figure} shows the qualitative results of ICDAR$2015$. The results indicate that our method can effectively handle multi-oriented text.
	
	\begin{table*}[t]
		\scriptsize
		\centering
		\caption{Scene text spotting results on Total-Text and CTW$1500$. “None” represents lexicon-free, while “Full” indicates all the words in the test set are used. $\P$ represents the speed on NVIDIA Tesla A$800$ GPUs.}
		\label{totaltext_ctw1500_table}
		\begin{tabular}{lccccccccccc}
			\toprule
			\multirow{3}{*}{Method}   & \multicolumn{5}{c}{Total-Text}   & \multicolumn{5}{c}{CTW$1500$}  & \multirow{3}{*}{FPS}  \\ \cmidrule{2-11}
			& \multicolumn{3}{c}{Detection} & \multicolumn{2}{c}{End-to-End} & \multicolumn{3}{c}{Detection} & \multicolumn{2}{c}{End-to-End} &                      \\ \cmidrule{2-11}
			& P        & R        & F       & None           & Full          & P        & R        & F       & None           & Full          &  \\ \midrule
			FOTS \cite{liu2018fots}            & $52.3$ & $38.0$ & $44.4$ & $32.2$ & -    & -    & -    & -    & $21.1$ & $39.7$ & -    \\
			Mask TextSpotter \cite{lyu2018mask}    &  $69.0$ & $55.0$ & $61.3$ & $52.9$ & $71.8$ & -    & -    & -    & -    & -    & $4.8$ \\
			TextDragon \cite{feng2019textdragon} & $85.6$ & $75.7$ & $80.3$ & $48.8$ & $74.8$ & $84.5$ & $82.8$ & $83.6$ & $39.7$ & $72.4$ & $2.6$  \\
			TextPerceptron \cite{qiao2020text}   & $88.8$ & $81.8$ & $85.2$ & $69.7$ & $78.3$ & $87.5$ & $81.9$ & $84.6$ & $57.0$ & -    & -    \\
			Mask TextSpotter v$3$ \cite{liao2020mask}    & -    & -    & -    & $71.2$ & $78.4$ & -    & -    & -    & -    & -    & -    \\
			ABCNet \cite{liu2020abcnet}   &-    & -    & -    & $64.2$ & $75.7$ & -    & -    & -    & $45.2$ & $74.1$ & $17.9$ \\
			PGNet \cite{wang2021pgnet}    & $85.5$ & $\mathbf{86.8}$ & $86.1$ & $63.1$ & -    & -    & -    & -    & -    & -    & $35.5$ \\
			MANGO \cite{qiao2021mango}    &-    & -    & -    & $72.9$ & $83.6$ & -    & -    & -    & $58.9$ & $78.7$ & $4.3$  \\
			ABCNetV$2$ \cite{liu2021abcnet} & $90.2$ & $84.1$ & $87.0$ & $70.4$ & $78.1$ & $85.6$ & $\mathbf{83.8}$ & $84.7$ & $57.5$ & $77.2$ & $10.0$ \\
			PAN++ \cite{wang2021pan++}    & -    & -    & -    & $68.6$ & $78.6$ & -    & -    & -    & -    & -    & $21.1$ \\
			TESTR \cite{zhang2022text}    & $93.4$ & $81.4$ & $86.9$ & $73.3$ & $83.9$ & $92.0$ & $82.6$ & $87.1$ & $56.0$ & $81.5$ & $5.3$  \\
			GLASS \cite{ronen2022glass}   & $90.8$ & $85.5$ & $\mathbf{88.1}$ & $\mathbf{79.9}$ & $86.2$ & -    & -    & -    & -    & -    & $3.0$  \\
			SwinTextSpotter \cite{huang2022swintextspotter} & -    & -    & $88.0$ & $74.3$ & $84.1$ & -    & -    & $88.0$ & $51.8$ & $77.0$ & -    \\
			TTS \cite{kittenplon2022towards}       & -    & -    & -    & $78.2$ & $86.3$ & -    & -    & -    & -    & -    & -    \\
			SPTS \cite{peng2022spts}            & -    & -    & -    & $74.2$ & $82.4$ & -    & -    & -    & $\mathbf{63.6}$ & $\mathbf{83.8}$ & -    \\
			\textbf{WordLenSpotter }    & $\mathbf{93.6}$ & $83.3$ & $\mathbf{88.1}$ & $76.5$ & $\mathbf{86.6}$ & $\mathbf{92.3}$ & $83.2$ & $87.5$ & $60.1$ & $79.7$ & $2.2\P$\\ \bottomrule
		\end{tabular}
	\end{table*}
	
	\begin{figure*}[t]
		\centering 
		\includegraphics[width=\linewidth]{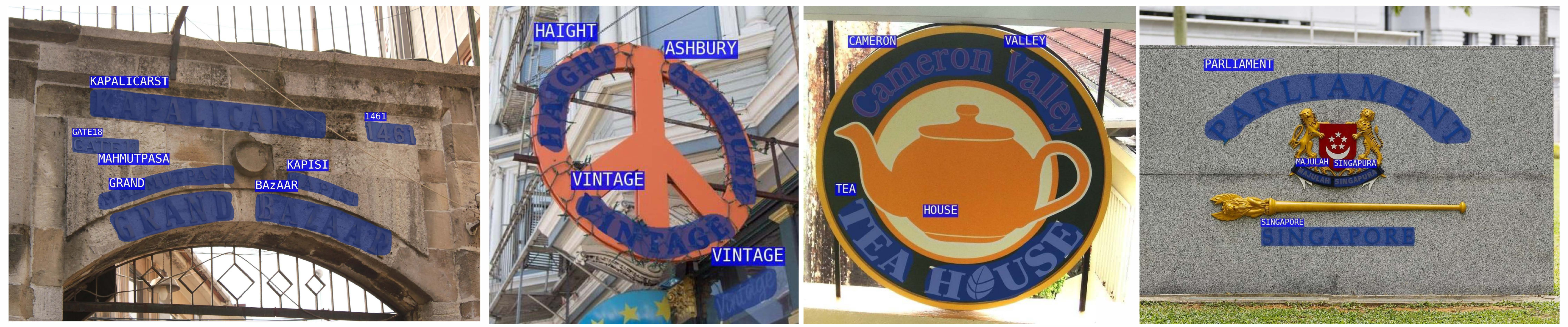}
		\caption{Qualitative results of WordLenSpotter on Total-Text.}
		\label{totaltext_vis_figure}
	\end{figure*}
	
	\begin{figure*}[t]
		\centering 
		\includegraphics[width=\linewidth]{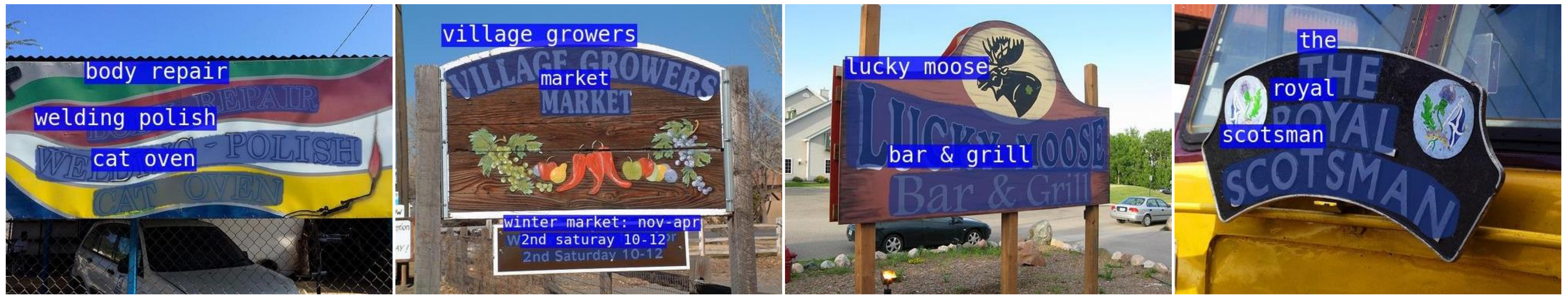}
		\caption{Qualitative results of WordLenSpotter on CTW$1500$.}
		\label{ctw1500_vis_figure}
	\end{figure*}
	
	We validate the effectiveness of WordLenSpotter for spotting arbitrary shaped text on two irregular text benchmarks, Total-Text and CTW$1500$, and the quantitative results are shown in Table \ref{totaltext_ctw1500_table}. In detection tasks, WordLenSpotter can achieve $88.1\%$ and $87.9\%$ F-scores on Total-Text and CTW$1500$, respectively. In the end-to-end text spotting task, WordLenSpotter achieves $86.6\%$ advanced performance when it has complete lexicons on Total-Text. Simultaneously, WordLenSpotter outperforms both SwinTextSpotter \cite{huang2022swintextspotter} and MANGO \cite{qiao2021mango} in terms of performance on the CTW$1500$ benchmark. Fig. \ref{totaltext_vis_figure} and \ref{ctw1500_vis_figure} show qualitative results for the two datasets. The results illustrate the strong capability of our method in accurately handling both straight and curved text. In summary, the performance evaluation on the Total-Text and CTW$1500$ benchmarks validates the effectiveness of our approach in effectively spotting text with arbitrary shapes.

	\subsection{Ablation study} \label{Section_4_4}
	
	\subsubsection{Number of text queries} \label{Section_4_4_1}
	
	The number of text instances in dense text scenes greatly exceeds those in regular scenarios, emphasizing the need to avoid setting the number of text queries too low, as it would lead to inadequate experimental results. To examine the impact of the number of text queries on WordLenSpotter performance in dense scenes, we conducted experiments with varying text query numbers on the DSTD$1500$ dataset. This allowed us to assess how the performance of WordLenSpotter is affected by the number of text queries in challenging scenarios with dense text instances. We train WordLenSpotter using $Q_{num}\in\{500,600,700,800\}$ as experimental conditions. The results in Table \ref{queries_num_table} indicate that the performance of WordLenSpotter in text detection tasks and end-to-end text spotting tasks improves with the increase in the number of text queries. And after $Q_{num}$ exceeds $700$, the performance of WordLenSpotter tends to saturate. Simultaneously, as the number of $Q_{num}$ increases, the computational cost will increase geometrically, which imposes strict requirements on computing resources. We balance the computational cost and performance of model training, and $Q_{num}=600$ is the most reasonable choice.
	
	\begin{table*}[tbp]
		\scriptsize
		\centering
		\caption{Experiment with the number of hyperparametric text queries $Q_{num}$ for WordLenSpotter on DSTD$1500$. “None” represents lexicon-free, while “Full” indicates all the words in the test set are used. GFlops represents the computational complexity of the model. Mem. represents the peak memory for batch processing of four images on an A$800$ GPU during training.}
		\label{queries_num_table}
		\begin{tabular}{cccccccc}
			\toprule
			\multirow{2}{*}{$Q_{num}$}  & \multicolumn{3}{c}{Detection} & \multicolumn{2}{c}{End-to-End} & \multirow{2}{*}{GFlops} & \multirow{2}{*}{Mem. (MB)} \\ \cmidrule{2-6} 
			& P        & R        & F       & None           & Full          \\ \midrule
			$500$ & $88.45$ & $78.01$ & $82.90$ & $75.10$ & $81.87$ & $\mathbf{221.8}$ & $\mathbf{44,740}$ \\
			$600$ & $88.64$ & $78.56$ & $83.30$ & $75.16$ & $81.95$ & $238.0$ & $53,863$ \\
			$700$ & $88.67$ & $78.60$ & $83.33$ & $75.20$ & $81.96$ & $264.2$ & $63,947$ \\
			$800$ & $\mathbf{88.68}$ & $\mathbf{78.60}$ & $\mathbf{83.34}$ & $\mathbf{75.21}$ & $\mathbf{81.97}$ & $308.6$ & $78,877$ \\ \bottomrule
		\end{tabular}
	\end{table*}
	
	\subsubsection{Architectural componentst} \label{Section_4_4_2}
	
	To evaluate the effectiveness of introducing components in WordLenSpotter for dense text spotting, we conduct ablation studies on DSTD$1500$. The SwinTextSpotter \cite{huang2022swintextspotter} architecture, which integrates text detection and recognition, serves as the baseline for our approach. To mitigate the influence of varying text query quantities, we specifically set the baseline text query quantity $Q_{num}$ to $600$. Subsequently, we conduct additional training and fine-tuning in the DSTD$1500$ dataset to acquire experimental data, which serves as the foundation for further analysis.
	
	\textbf{BiFPN-DC.} As shown in Table \ref{ablation_table}, the image encoder adopts an efficient bidirectional cross-scale connection and weighted feature fusion network with dilated convolution, the results of text detection in dense scenes are improved by $0.47\%$. For the end-to-end text spotting task, the spotting results improve by $0.53\%$ and $0.39\%$ without and with the lexicon, respectively. This is mainly because BiFPN with dilated convolution can effectively fuse multi-scale features extracted by Swin Transformer. Additionally, it also increases the receptive field of the network, which is conducive to the discovery of text instances with various scales.
	
	\begin{table*}[t]
		\scriptsize
		\centering
		\caption{Ablation studies of WordLenSpotter. \textbf{BiFPN-DC} represents the use of an efficient bidirectional cross-scale connection and weighted feature fusion network with dilated convolution. \textbf{SLP} represents the introduction of the Spatial Length Predictor. \textbf{LenSeg Head} represents the introduction of the word Length-based Segmentation Head. The end-to-end positioning metric “None” (“Full”) indicates that the F-score of the lexicon containing all words in the test set is not used.}
		\label{ablation_table}
		\begin{tabular}{cccccccc}
			\toprule
			\multirow{2}{*}{BiFPN-DC} & \multirow{2}{*}{SLP} & \multirow{2}{*}{LenSeg Head} & \multicolumn{3}{c}{Detection} & \multicolumn{2}{c}{End-to-End} \\ \cmidrule{4-8} 
			&                        &                      & P        & R        & F       & None           & Full          \\ \midrule
			- & - & - & $86.01$    & $78.00$    & $81.81$   & $74.57$          & $81.45$         \\
			\checkmark    & - & - & $86.37$    & $78.56$    & $82.28$  & $75.10$          & $81.84$         \\
			\checkmark   & \checkmark  & -  & $86.51$    & $78.49$     & $82.31$   & $75.10$          & $81.91$        \\
			\checkmark   & - & \checkmark   & $86.45$    & $\mathbf{78.65}$    & $82.37$   & $75.07$          & $81.86$ \\
			- & \checkmark   & \checkmark  & $87.05$    & $78.50$    & $82.55$   & $74.77$     & $81.71$         \\
			\checkmark  & \checkmark   & \checkmark    & $\mathbf{88.64}$    & $78.57$    & $\mathbf{83.30}$   & $\mathbf{75.16}$          & $\mathbf{81.95}$         \\ \bottomrule
		\end{tabular}
	\end{table*}
	
	\textbf{SLP.} The introduction of the text word length information prediction branch improves the detection task results by $0.50\%$. The end-to-end text spotting task has an improvement of $0.53\%$ and $0.46\%$ in spotting results compared to a baseline without and with the lexicon, respectively. This is attributed to the direct enhancement of the feature learning capability of text queries for different text instance shapes and character counts in dense scene images by the Spatial Length Predictor. It facilitates the synergistic collaboration between detection and recognition, resulting in the accurate matching of text instances.
	
	\textbf{LenSeg Head.} We evaluate the effectiveness of the word Length-aware Segmentation Head. In Table \ref{ablation_table}, the introduction of the word length-aware segmentation task significantly improves the recall rate of detection tasks, while also increasing the F-score by $0.56\%$. And for the end-to-end text spotting task, the spotting results are improved by $0.50\%$ and $0.41\%$ compared to a baseline without and with the lexicon, respectively. The experimental results demonstrate that the LenSeg Head enhances the perception of text queries regarding the precise location of text instances with extreme word length in dense scene text.
	
	We further evaluate the performance of the proposed SLP and LenSeg Head on DSTD$1500$ without utilizing the BiFPN-DC in the image encoder. As shown in Table \ref{ablation_table}, compared to the baseline, this improves the results of the detection task by $0.74\%$, reaching $82.55\%$. For the end-to-end text spotting task, the spotting results achieve $74.77\%$ and $81.71\%$ without and with the lexicon, respectively. The results indicate that our proposed tasks of text word length prediction and word length-aware segmentation effectively enhance the learning of model decoders within conventional text detection and recognition collaborative architectures that rely on text queries.
	
	\section{Conclusion} \label{Section_5}
	
	In this paper, we present WordLenSpotter, a novel method for text spotting in dense scenes, built upon the collaborative architecture of text detection and recognition. Our approach specifically leverages the prior knowledge of a strong correlation between text instance aspect ratio and character quantity. Our image encoder adopts Swin Transformer to model the global context of text images in dense scenes and adopts an efficient bidirectional cross-scale connection and weighted feature fusion network equipped with dilated convolution to fuse multiscale image features. Furthermore, we integrate a text Spatial Length Predictor and a word Length-aware Segmentation Head, significantly improving the ability of text queries to accurately perceive both short and long words within the long-tailed distribution of clearly visible text instances in dense scenes. In addition, we introduce the DSTD$1500$ dataset, a real-world dataset specifically designed for dense scene text. Through extensive experimentation on both DSTD$1500$ and publicly available benchmarks, we demonstrate that WordLenSpotter achieves state-of-the-art performance in end-to-end text spotting, and effectively handles extreme word length text instances in dense scenes with long-tailed distributions of word length.
	
	Considering our algorithm requirements, the process of generating self-generated labels for certain training samples can potentially affect training efficiency. For instance, generating word-length information labels for dense samples containing curved text instances can take tens of times longer compared to sparse samples annotated with quadrilaterals. In future research, we aim to optimize the label generation method while ensuring a clear separation between the label generation process and the training process, thereby improving overall efficiency.
	
	\section*{Acknowledgment}
	This work was supported in part by the National Natural Science Foundation of China under Grant $62171327$, $62171328$ and $62072350$, and in part by Key R\&D Program in Hubei Province, China under Grant $2022$BAA$079$.

	\ifCLASSOPTIONcaptionsoff
	\newpage
	\fi
	
	\balance
	\bibliography{cas-refs-my}{}

\begin{thebibliography}{10}
\providecommand{\url}[1]{#1}
\csname url@samestyle\endcsname
\providecommand{\newblock}{\relax}
\providecommand{\bibinfo}[2]{#2}
\providecommand{\BIBentrySTDinterwordspacing}{\spaceskip=0pt\relax}
\providecommand{\BIBentryALTinterwordstretchfactor}{4}
\providecommand{\BIBentryALTinterwordspacing}{\spaceskip=\fontdimen2\font plus
\BIBentryALTinterwordstretchfactor\fontdimen3\font minus
  \fontdimen4\font\relax}
\providecommand{\BIBforeignlanguage}[2]{{%
\expandafter\ifx\csname l@#1\endcsname\relax
\typeout{** WARNING: IEEEtran.bst: No hyphenation pattern has been}%
\typeout{** loaded for the language `#1'. Using the pattern for}%
\typeout{** the default language instead.}%
\else
\language=\csname l@#1\endcsname
\fi
#2}}
\providecommand{\BIBdecl}{\relax}
\BIBdecl

\bibitem{tian2016detecting}
Z.~Tian, W.~Huang, T.~He, P.~He, and Y.~Qiao, ``Detecting text in natural image
  with connectionist text proposal network,'' in \emph{Proceedings of the
  European Conference on Computer Vision}, 2016, pp. 56--72.

\bibitem{liao2018textboxes++}
M.~Liao, B.~Shi, and X.~Bai, ``Textboxes++: A single-shot oriented scene text
  detector,'' \emph{IEEE Transactions on Image Processing}, vol.~27, no.~8, pp.
  3676--3690, 2018.

\bibitem{zhou2017east}
X.~Zhou, C.~Yao, H.~Wen, Y.~Wang, S.~Zhou, W.~He, and J.~Liang, ``East: an
  efficient and accurate scene text detector,'' in \emph{Proceedings of the
  IEEE Conference on Computer Vision and Pattern Recognition}, 2017, pp.
  5551--5560.

\bibitem{shi2016end}
B.~Shi, X.~Bai, and C.~Yao, ``An end-to-end trainable neural network for
  image-based sequence recognition and its application to scene text
  recognition,'' \emph{IEEE Transactions on Pattern Analysis and Machine
  Intelligence}, vol.~39, no.~11, pp. 2298--2304, 2017.

\bibitem{li2017towards}
H.~Li, P.~Wang, and C.~Shen, ``Towards end-to-end text spotting with
  convolutional recurrent neural networks,'' in \emph{Proceedings of the IEEE
  International Conference on Computer Vision}, 2017, pp. 5238--5246.

\bibitem{liu2018fots}
X.~Liu, D.~Liang, S.~Yan, D.~Chen, Y.~Qiao, and J.~Yan, ``Fots: Fast oriented
  text spotting with a unified network,'' in \emph{Proceedings of the IEEE
  Conference on Computer Vision and Pattern Recognition}, 2018, pp. 5676--5685.

\bibitem{wang2021pan++}
W.~Wang, E.~Xie, X.~Li, X.~Liu, D.~Liang, Z.~Yang, T.~Lu, and C.~Shen, ``Pan++:
  Towards efficient and accurate end-to-end spotting of arbitrarily-shaped
  text,'' \emph{IEEE Transactions on Pattern Analysis and Machine
  Intelligence}, vol.~44, no.~9, pp. 5349--5367, 2022.

\bibitem{huang2022swintextspotter}
M.~Huang, Y.~Liu, Z.~Peng, C.~Liu, D.~Lin, S.~Zhu, N.~Yuan, K.~Ding, and
  L.~Jin, ``Swintextspotter: Scene text spotting via better synergy between
  text detection and text recognition,'' in \emph{Proceedings of the IEEE
  Conference on Computer Vision and Pattern Recognition}, 2022, pp. 4593--4603.

\bibitem{zhang2022text}
X.~Zhang, Y.~Su, S.~Tripathi, and Z.~Tu, ``Text spotting transformers,'' in
  \emph{Proceedings of the IEEE Conference on Computer Vision and Pattern
  Recognition}, 2022, pp. 9519--9528.

\bibitem{peng2022spts}
D.~Peng, X.~Wang, Y.~Liu, J.~Zhang, M.~Huang, S.~Lai, J.~Li, S.~Zhu, D.~Lin,
  C.~Shen \emph{et~al.}, ``Spts: single-point text spotting,'' in
  \emph{Proceedings of the 30th ACM International Conference on Multimedia},
  2022, pp. 4272--4281.

\bibitem{vaswani2017attention}
A.~Vaswani, N.~Shazeer, N.~Parmar, J.~Uszkoreit, L.~Jones, A.~N. Gomez,
  {\L}.~Kaiser, and I.~Polosukhin, ``Attention is all you need,''
  \emph{Advances in Neural Information Processing Systems}, vol.~30, 2017.

\bibitem{karatzas2015icdar}
D.~Karatzas, L.~Gomez-Bigorda, A.~Nicolaou, S.~Ghosh, A.~Bagdanov, M.~Iwamura,
  J.~Matas, L.~Neumann, V.~R. Chandrasekhar, S.~Lu \emph{et~al.}, ``Icdar 2015
  competition on robust reading,'' in \emph{2015 13th International Conference
  on Document Analysis and Recognition}.\hskip 1em plus 0.5em minus 0.4em\relax
  IEEE, 2015, pp. 1156--1160.

\bibitem{ch2020total}
C.-K. Ch’ng, C.~S. Chan, and C.-L. Liu, ``Total-text: toward orientation
  robustness in scene text detection,'' \emph{International Journal on Document
  Analysis and Recognition}, vol.~23, no.~1, pp. 31--52, 2020.

\bibitem{liu2019curved}
Y.~Liu, L.~Jin, S.~Zhang, C.~Luo, and S.~Zhang, ``Curved scene text detection
  via transverse and longitudinal sequence connection,'' \emph{Pattern
  Recognition}, vol.~90, pp. 337--345, 2019.

\bibitem{bissacco2013photoocr}
A.~Bissacco, M.~Cummins, Y.~Netzer, and H.~Neven, ``Photoocr: Reading text in
  uncontrolled conditions,'' in \emph{Proceedings of the IEEE International
  Conference on Computer Vision}, 2013, pp. 785--792.

\bibitem{liao2017textboxes}
M.~Liao, B.~Shi, X.~Bai, X.~Wang, and W.~Liu, ``Textboxes: A fast text detector
  with a single deep neural network,'' in \emph{Proceedings of the AAAI
  Conference on Artificial Intelligence}, vol.~31, no.~1, 2017.

\bibitem{baek2019character}
Y.~Baek, B.~Lee, D.~Han, S.~Yun, and H.~Lee, ``Character region awareness for
  text detection,'' in \emph{Proceedings of the IEEE Conference on Computer
  Vision and Pattern Recognition}, 2019, pp. 9365--9374.

\bibitem{wang2019arbitrary}
X.~Wang, Y.~Jiang, Z.~Luo, C.-L. Liu, H.~Choi, and S.~Kim, ``Arbitrary shape
  scene text detection with adaptive text region representation,'' in
  \emph{Proceedings of the IEEE Conference on Computer Vision and Pattern
  Recognition}, 2019, pp. 6449--6458.

\bibitem{yu2023turning}
W.~Yu, Y.~Liu, W.~Hua, D.~Jiang, B.~Ren, and X.~Bai, ``Turning a clip model
  into a scene text detector,'' in \emph{Proceedings of the IEEE Conference on
  Computer Vision and Pattern Recognition}, 2023, pp. 6978--6988.

\bibitem{yao2014strokelets}
C.~Yao, X.~Bai, B.~Shi, and W.~Liu, ``Strokelets: A learned multi-scale
  representation for scene text recognition,'' in \emph{Proceedings of the IEEE
  Conference on Computer Vision and Pattern Recognition}, 2014, pp. 4042--4049.

\bibitem{su2017accurate}
B.~Su and S.~Lu, ``Accurate recognition of words in scenes without character
  segmentation using recurrent neural network,'' \emph{Pattern Recognition},
  vol.~63, pp. 397--405, 2017.

\bibitem{lee2016recursive}
C.-Y. Lee and S.~Osindero, ``Recursive recurrent nets with attention modeling
  for ocr in the wild,'' in \emph{Proceedings of the IEEE Conference on
  Computer Vision and Pattern Recognition}, 2016, pp. 2231--2239.

\bibitem{luo2019moran}
C.~Luo, L.~Jin, and Z.~Sun, ``Moran: A multi-object rectified attention network
  for scene text recognition,'' \emph{Pattern Recognition}, vol.~90, pp.
  109--118, 2019.

\bibitem{liu2023towards}
C.~Liu, C.~Yang, H.-B. Qin, X.~Zhu, C.-L. Liu, and X.-C. Yin, ``Towards
  open-set text recognition via label-to-prototype learning,'' \emph{Pattern
  Recognition}, vol. 134, p. 109109, 2023.

\bibitem{lyu2018mask}
P.~Lyu, M.~Liao, C.~Yao, W.~Wu, and X.~Bai, ``Mask textspotter: An end-to-end
  trainable neural network for spotting text with arbitrary shapes,'' in
  \emph{Proceedings of the European Conference on Computer Vision}, 2018, pp.
  67--83.

\bibitem{liao2021mask}
M.~Liao, P.~Lyu, M.~He, C.~Yao, W.~Wu, and X.~Bai, ``Mask textspotter: An
  end-to-end trainable neural network for spotting text with arbitrary
  shapes,'' \emph{IEEE Transactions on Pattern Analysis and Machine
  Intelligence}, vol.~43, no.~2, pp. 532--548, 2021.

\bibitem{liao2020mask}
M.~Liao, G.~Pang, J.~Huang, T.~Hassner, and X.~Bai, ``Mask textspotter v3:
  Segmentation proposal network for robust scene text spotting,'' in
  \emph{Proceedings of the European Conference on Computer Vision}, 2020, pp.
  706--722.

\bibitem{feng2019textdragon}
W.~Feng, W.~He, F.~Yin, X.-Y. Zhang, and C.-L. Liu, ``Textdragon: An end-to-end
  framework for arbitrary shaped text spotting,'' in \emph{Proceedings of the
  IEEE International Conference on Computer Vision}, 2019, pp. 9076--9085.

\bibitem{qiao2020text}
L.~Qiao, S.~Tang, Z.~Cheng, Y.~Xu, Y.~Niu, S.~Pu, and F.~Wu, ``Text perceptron:
  Towards end-to-end arbitrary-shaped text spotting,'' in \emph{Proceedings of
  the AAAI Conference on Artificial Intelligence}, vol.~34, no.~07, 2020, pp.
  11\,899--11\,907.

\bibitem{ronen2022glass}
R.~Ronen, S.~Tsiper, O.~Anschel, I.~Lavi, A.~Markovitz, and R.~Manmatha,
  ``Glass: Global to local attention for scene-text spotting,'' in
  \emph{Proceedings of the European Conference on Computer Vision}, 2022, pp.
  249--266.

\bibitem{wang2021pgnet}
P.~Wang, C.~Zhang, F.~Qi, S.~Liu, X.~Zhang, P.~Lyu, J.~Han, J.~Liu, E.~Ding,
  and G.~Shi, ``Pgnet: Real-time arbitrarily-shaped text spotting with point
  gathering network,'' in \emph{Proceedings of the AAAI Conference on
  Artificial Intelligence}, vol.~35, no.~4, 2021, pp. 2782--2790.

\bibitem{qiao2021mango}
L.~Qiao, Y.~Chen, Z.~Cheng, Y.~Xu, Y.~Niu, S.~Pu, and F.~Wu, ``Mango: A mask
  attention guided one-stage scene text spotter,'' in \emph{Proceedings of the
  AAAI Conference on Artificial Intelligence}, vol.~35, no.~3, 2021, pp.
  2467--2476.

\bibitem{liu2020abcnet}
Y.~Liu, H.~Chen, C.~Shen, T.~He, L.~Jin, and L.~Wang, ``Abcnet: Real-time scene
  text spotting with adaptive bezier-curve network,'' in \emph{Proceedings of
  the IEEE Conference on Computer Vision and Pattern Recognition}, 2020, pp.
  9809--9818.

\bibitem{liu2021abcnet}
Y.~Liu, C.~Shen, L.~Jin, T.~He, P.~Chen, C.~Liu, and H.~Chen, ``Abcnet v2:
  Adaptive bezier-curve network for real-time end-to-end text spotting,''
  \emph{IEEE Transactions on Pattern Analysis and Machine Intelligence},
  vol.~44, no.~11, pp. 8048--8064, 2022.

\bibitem{chen2021pix2seq}
T.~Chen, S.~Saxena, L.~Li, D.~J. Fleet, and G.~Hinton, ``Pix2seq: A language
  modeling framework for object detection,'' \emph{arXiv preprint
  arXiv:2109.10852}, 2021.

\bibitem{carion2020end}
N.~Carion, F.~Massa, G.~Synnaeve, N.~Usunier, A.~Kirillov, and S.~Zagoruyko,
  ``End-to-end object detection with transformers,'' in \emph{Proceedings of
  the European Conference on Computer Vision}, 2020, pp. 213--229.

\bibitem{kittenplon2022towards}
Y.~Kittenplon, I.~Lavi, S.~Fogel, Y.~Bar, R.~Manmatha, and P.~Perona, ``Towards
  weakly-supervised text spotting using a multi-task transformer,'' in
  \emph{Proceedings of the IEEE Conference on Computer Vision and Pattern
  Recognition}, 2022, pp. 4604--4613.

\bibitem{liu2021swin}
Z.~Liu, Y.~Lin, Y.~Cao, H.~Hu, Y.~Wei, Z.~Zhang, S.~Lin, and B.~Guo, ``Swin
  transformer: Hierarchical vision transformer using shifted windows,'' in
  \emph{Proceedings of the IEEE International Conference on Computer Vision},
  2021, pp. 10\,012--10\,022.

\bibitem{tan2020efficientdet}
M.~Tan, R.~Pang, and Q.~V. Le, ``Efficientdet: Scalable and efficient object
  detection,'' in \emph{Proceedings of the IEEE Conference on Computer Vision
  and Pattern Recognition}, 2020, pp. 10\,781--10\,790.

\bibitem{rezatofighi2019generalized}
H.~Rezatofighi, N.~Tsoi, J.~Gwak, A.~Sadeghian, I.~Reid, and S.~Savarese,
  ``Generalized intersection over union: A metric and a loss for bounding box
  regression,'' in \emph{Proceedings of the IEEE Conference on Computer Vision
  and Pattern Recognition}, 2019, pp. 658--666.

\bibitem{milletari2016v}
F.~Milletari, N.~Navab, and S.-A. Ahmadi, ``V-net: Fully convolutional neural
  networks for volumetric medical image segmentation,'' in \emph{2016 Fourth
  International Conference on 3D Vision}.\hskip 1em plus 0.5em minus
  0.4em\relax IEEE, 2016, pp. 565--571.

\bibitem{pratikakis2013icdar}
I.~Pratikakis, B.~Gatos, and K.~Ntirogiannis, ``Icdar 2013 document image
  binarization contest (dibco 2013),'' in \emph{2013 12th International
  Conference on Document Analysis and Recognition}.\hskip 1em plus 0.5em minus
  0.4em\relax IEEE, 2013, pp. 1471--1476.

\bibitem{nayef2017icdar2017}
N.~Nayef, F.~Yin, I.~Bizid, H.~Choi, Y.~Feng, D.~Karatzas, Z.~Luo, U.~Pal,
  C.~Rigaud, J.~Chazalon \emph{et~al.}, ``Icdar2017 robust reading challenge on
  multi-lingual scene text detection and script identification-rrc-mlt,'' in
  \emph{2017 14th IAPR International Conference on Document Analysis and
  Recognition}, vol.~1.\hskip 1em plus 0.5em minus 0.4em\relax IEEE, 2017, pp.
  1454--1459.

\bibitem{sun2021sparse}
P.~Sun, R.~Zhang, Y.~Jiang, T.~Kong, C.~Xu, W.~Zhan, M.~Tomizuka, L.~Li,
  Z.~Yuan, C.~Wang \emph{et~al.}, ``Sparse r-cnn: End-to-end object detection
  with learnable proposals,'' in \emph{Proceedings of the IEEE/CVF conference
  on computer vision and pattern recognition}, 2021, pp. 14\,454--14\,463.

\end{thebibliography}
	\bibliographystyle{IEEEtran}
	
\end{document}